\documentclass[conference]{IEEEtran}
\usepackage{times}

\usepackage[numbers]{natbib}
\usepackage{multicol}
\usepackage[bookmarks=true]{hyperref}
\usepackage{xcolor}
\usepackage{amsmath}
\usepackage{amssymb}
\usepackage{bbm}
\usepackage{graphicx}
\usepackage{subfig}
\usepackage{booktabs}
\usepackage{multirow}

\begin{document}

\newcommand{\shrink}{\def\baselinestretch{0.95}\large\normalsize} 
\shrink

\title{Human Preference Modeling Using Visual Motion Prediction Improves Robot Skill Learning from Egocentric Human Video}


\author{\authorblockN{Mrinal Verghese}
\authorblockA{Robotics Institute\
Carnegie Mellon University\\
Email: mverghes@andrew.cmu.edu}
\and
\authorblockN{Christopher G Atkeson}
\authorblockA{Robotics Institute\
Carnegie Mellon University\\
Email: cga@andrew.cmu.edu}}


%

\maketitle

\begin{abstract}
We present an approach to robot learning from egocentric human videos by modeling human preferences in a reward function and optimizing robot behavior to maximize this reward. Prior work on reward learning from human videos attempts to measure the long-term value of a visual state as the temporal distance between it and the terminal state in a demonstration video. These approaches make assumptions that limit performance when learning from video. They must also transfer the learned value function across the embodiment and environment gap. Our method models human preferences by learning to predict the motion of tracked points between subsequent images and defines a reward function as the agreement between predicted and observed object motion in a robot's behavior at each step. We then use a modified Soft Actor Critic (SAC) algorithm initialized with 10 on-robot demonstrations to estimate a value function from this reward and optimize a policy that maximizes this value function, all on the robot. Our approach is capable of learning on a real robot, and we show that policies learned with our reward model match or outperform prior work across multiple tasks in both simulation and on the real robot.
\end{abstract}

\IEEEpeerreviewmaketitle

\section{Introduction}




How can we dramatically reduce the number of on-robot demonstrations needed to learn basic skills by transferring information from human video demonstrations of skills to robots learning these same skills? When training robots with teleoperation demonstrations collected on the robot, there often exists a ``skill-gap'' where the learned policy underperforms the demonstrations. This gap exists because robot teleoperation demonstrations are expensive to collect and often only cover a portion of the task space. Videos of humans performing household tasks are a readily available source of supervision. These videos can help close this gap by covering more of the task space, but robots face challenges in learning from this data due to mismatches in the actor embodiment and the environment. A very promising approach to leverage human video for robot skill learning is to learn a feedback signal, such as a reward or cost function, from large video datasets, and then optimize this function on the robot via planning or reinforcement learning. This approach helps mitigate the embodiment gap and does not require tracking and imitating human hand or object trajectories, methods that can struggle with contact-rich tasks or require large amounts of on-robot data. The most common strategy to learn these feedback signals is to estimate the long-term value $V(s)$ of a state as its temporal distance to the goal (the last frame in a video demonstration) by assuming every frame in the demonstration incurs a fixed per-frame cost and assigning the value of a visual state as its frame distance to the goal~\citep{liuTimeRewarderLearningDense2025, maVIPUniversalVisual2023, maLIVLanguageImageRepresentations2023b, sermanetTimeContrastiveNetworksSelfSupervised2018a, milikicVLDVisualLanguage2025}. However, this strategy is susceptible to noise in the demonstration data and can have trouble generalizing this learned value function to robot videos and recognizing failed attempts. In this work, we present a novel method of learning reward functions from large human video datasets and argue that representing human preferences via visual motion features is a more effective way to learn a reward function from human video demonstrations.


There are several challenges associated with learning robot reward signals from large, unstructured human video datasets. These datasets generally only contain successful demonstrations, and methods that do not select the right features to attend to may falsely reward episodes that look similar to the demonstrations but do not achieve the same effect. Humans often multitask, and video datasets contain many instances where a human may interrupt one skill to perform another, hesitate during a skill, or otherwise not act in a time-optimal manner. Methods that measure value by temporal distance to the goal will negatively bias all states before any such interruption. Finally, there is a large distribution shift between human video and robot video, and whatever quantity that is learned from human video must be transferred across this gap in a robust way.

In this work, we approach reward learning as a problem of modeling the short-term preferences of human demonstrators at each step in a demonstration. We learn models to predict how tracked points on task-relevant objects in a demonstration video move at each step, and assign a per-transition reward to a robot rollout based on the alignment between the predicted point motion and the observed point motion induced by the robot's actions. We then utilize a modified version of the Soft Actor Critic (SAC) algorithm to estimate the long-term value of a state in the robot's state distribution and optimize a policy to maximize this value. By predicting point motion of task-relevant objects, and rewarding the robot for matching this point motion, we explicitly encourage the robot to match the outcomes of the human demonstrations and not just the human's actions. By modeling short-term preferences rather than long-term value estimates, our approach is less susceptible to bias from hesitation or distractions in the demonstrations or from distribution shift. 

\begin{figure*}[t] 
    \centering

    \includegraphics[width=0.99\linewidth, trim={0cm 8cm 0cm 0cm},clip]{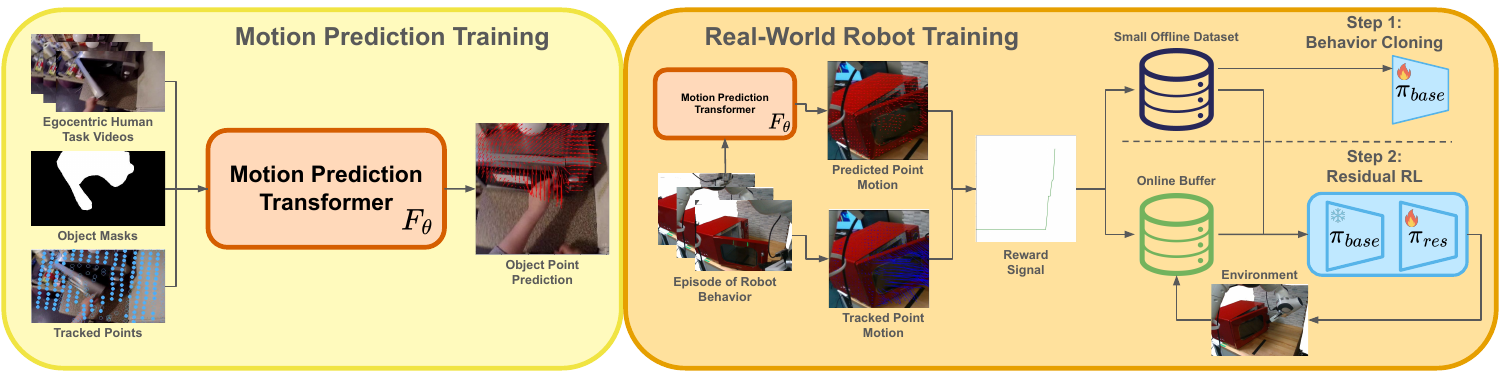}

  \caption{\textbf{Modeling human preferences using Motion Prediction Reward improves robot skill learning from human video data.} To learn a robust reward signal from human video for a given task $\mathcal{T}$ (in this case, ``Open Microwave''), we first extract point tracks and object masks using off-the-shelf models from a set of egocentric human videos demonstrating $\mathcal{T}$. This data is used to train our Motion Prediction Transformer ($F_\theta$) that can predict how points on an object will move given a visual observation. This model is used to calculate reward for an episode of robot behavior, by tracking points in a video of the episode, and measuring the alignment between predicted and observed point motion at each transition. To learn a robot policy for task $\mathcal{T}$, we collect a very small set of demonstrations (small offline dataset) and train a behavior cloning policy ($\pi_{base}$) to kickstart learning. We then use a sample-efficient residual RL framework that leverages both the collected demonstrations and new episodes from online interaction (online buffer), labeled with our reward model, to improve its performance on the given task. This process is able to increase a robot's success rate on a task by over 30\% with just an hour of real-world interaction, and significantly outperforms prior work on reward learning from human video.}
  \label{fig:Intro} 
  \vspace{-3mm}
\end{figure*}

To improve a behavior-cloned robot policy's performance on a specific task, our method first analyzes a set of egocentric videos of humans performing that task, and outputs a preference model that predicts how a set of tracked points across task-relevant objects will move between frames. For each attempt the robot makes to complete the task, we compare the predicted point motion at each transition with the observed point motion to generate a reward signal. We then use an actor-critic reinforcement learning algorithm in a residual RL framework that learns a value function and policy corrections on top of the behavior-cloned policy to make the robot's behavior more closely match the predicted human preferences. This pipeline is highlighted in Figure \ref{fig:Intro}. This RL framework combines modern advances in imitation learning, such as diffusion policies, with sample-efficient reinforcement learning techniques to enable practical online finetuning of robot policies in the real world. We demonstrate the performance of this approach in a real-world reinforcement learning setting across three tasks using our learned reward models, and show that our method can enable a robot to rapidly improve its own performance in only an hour of wall-clock training time.

\section{Related Works}

\subsection{Robot Learning with Human Video Data}
Human video data exists in large quantities on the internet, is easy to collect, and has the potential to greatly reduce the number of on-robot demonstrations needed to learn new robot skills. To accomplish this, prior work has explored a wide variety of approaches to transfer information from this data source to robot policies. Several works have explored objectives like contrastive learning~\citep{nairR3MUniversalVisual2022d} and masked autoencoding~\citep{majumdarWhereAreWe2024a} to learn general-purpose visual encoders from common egocentric video datasets such as Ego4D~\citep{graumanEgo4DWorld30002022a} and Epic Kitchens~\citep{damenScalingEgocentricVision2018a, damenEPICKITCHENSDatasetCollection2020}. These vision encoders have primarily been used to reduce the visual learning burden when learning robot policies, but are also capable of measuring similarity between states to determine reward. Recently, the popular Vision-Language-Action (VLA) architecture has been adapted to use video generation models trained on human videos as their pretrained backbone~\citep{shenVideoVLAVideoGenerators2025, kimCosmosPolicyFineTuning2026}. These Video VLA models show promising improvements over existing VLAs, but also suffer from some of the drawbacks of VLAs, including being expensive to train and finetune. Other works have trained world models using human video data that are capable of forecasting actions into the future~\citep{kimCosmosPolicyFineTuning2026, assranVJEPA2SelfSupervised2025, wangLatentPolicySteering2025}. Finally, a large body of work has examined directly translating human videos into robot behavior. The approaches can be broken down into three main categories. The first category attempts to directly mimic or forecast the trajectory of objects in a video and then solve for a set of robot actions to match that object trajectory in the real world~\citep{heppertDITTODemonstrationImitation2024, paloDINOBotRobotManipulation2024, bharadhwajTrack2ActPredictingPoint2024a, liOKAMITeachingHumanoid2024}. Some of these methods can learn from a single video demonstration, but as they compute trajectories and not closed-loop policies, they tend to be less robust and struggle with contact-rich tasks. Several of these works do collect extra robot data to mitigate this limitation~\citep{bharadhwajTrack2ActPredictingPoint2024a, liOKAMITeachingHumanoid2024}. The next category is methods that perform imitation learning on human videos by tracking human hands and retargeting the human hand actions to a robot's action space~\citep{qinDexMVImitationLearning2022, bahetyScrewMimicBimanualImitation2024a, kareerEgoMimicScalingImitation2024, liuEgoZeroRobotLearning2025, renMotionTracksUnified2025, lepertMasqueradeLearningInthewild2025}. These methods show impressive generalization, but depend heavily on how the human-to-robot retargeting is done and usually require either in-domain videos (as opposed to internet videos) in the target environment or reasonable amounts of robot data. Finally, recent work by \citet{yuanHERMESHumantoRobotEmbodied2025} shows impressive real-world results by leveraging reinforcement learning to better track trajectories from videos. However, this approach requires recreating both the robot and environment in simulation and transferring the policy learned in simulation back to the real world. This requires significant engineering effort and can be challenging for tasks involving deformable objects.


\subsection{Learning Visual Reward Models in Robotics}
In this work, we address the problem of learning reward models that operate on visual inputs. Estimating rewards from visual input enables reward learning from diverse sources, including pretrained image~\citep{mahmoudiehZeroShotRewardSpecification2022a} and video~\citep{shaoConcept2RobotLearningManipulation} models, diverse robot datasets~\citep{leeRoboRewardGeneralPurposeVisionLanguage2026}, robot video~\citep{liuTimeRewarderLearningDense2025, ghasemipourSelfImprovingEmbodiedFoundation2025, mandikalMashSpreadSlice2025}, in-domain human video demonstrations~\citep{liuTimeRewarderLearningDense2025, sermanetTimeContrastiveNetworksSelfSupervised2018a, guzeyBridgingHumanRobot2024}, and large video datasets~\citep{milikicVLDVisualLanguage2025, maVIPUniversalVisual2023, maLIVLanguageImageRepresentations2023b}. Most of these approaches learn an encoder that maps visual state into a latent space, and measures the distance between a visual observation and a goal input in that latent space to determine the value of a visual state. For the visual reward learning methods that leverage video demonstrations, they primarily train these encoders to estimate the temporal distance between a visual state and a goal, as measured by the number of frames between them~\citep{liuTimeRewarderLearningDense2025, maVIPUniversalVisual2023, maLIVLanguageImageRepresentations2023b, sermanetTimeContrastiveNetworksSelfSupervised2018a, milikicVLDVisualLanguage2025}. While this metric is convenient for video data and has demonstrated success in the past, it can be less effective for training on large-scale video datasets due to the issues highlighted above. We compare our approach to VIP from \citet{maVIPUniversalVisual2023} due to its similar training data (large-scale egocentric human video datasets), and available pretrained models. It's important to note that our reward model can easily be composed with other learned reward signals. A practical deployment of reward models for robot learning might combine an approach like ours that provides dense, per-step feedback, with an approach like RoboReward from \citet{leeRoboRewardGeneralPurposeVisionLanguage2026} that provides a high-level success indication at the end of an episode. Finally, there is also a large body of work in Inverse Reinforcement Learning, which learns reward functions from expert demonstrations on the robot, but this work often operates under a very different set of assumptions from visual reward learning.



\subsection{Robot Learning from Demonstration and Practice}
To enable our robot to improve its performance with our reward signal, we turn to the field of learning from demonstration and practice. Reinforcement Learning (RL) is a powerful method for optimizing robot behavior according to a reward or cost function, but it is also notoriously sample inefficient, particularly in its initial exploration phase. Combining RL with a small set of on-robot demonstrations helps shortcut this sample inefficient exploration phase and greatly accelerates the learning process. Some prior work in this area used parameterized primitives to simplify the action space and enable efficient online adaptation~\cite{bentivegnaLearningObservationPractice, yamaguchiLearningPouringSkills2014}. A large body of recent work has explored adapting behavior cloned neural network policies with reinforcement learning on real robot hardware~\citep{ankileImitationRefinementResidual2024a, ankileResidualOffPolicyRL2025, luoSERLSoftwareSuite2025, luoPreciseDexterousRobotic2025, jainSmoothSeaNever2025, huImitationBootstrappedReinforcement2024, markPolicyAgnosticRL2024,ballEfficientOnlineReinforcement2023, renDiffusionPolicyPolicy2024}. The primary challenge in this area is to make an algorithm that is both stable and sample-efficient. Some works have tackled this by identifying design decisions to make existing Off-Policy RL algorithms like Soft Actor Critic (SAC)~\citep{guptaSoftActorCritic} work nicely for real-world learning~\citep{ballEfficientOnlineReinforcement2023, luoSERLSoftwareSuite2025, luoPreciseDexterousRobotic2025}. Other works have eschewed typical RL training objectives like policy gradients or Q-function maximization for supervised learning, which is typically more stable~\citep{jainSmoothSeaNever2025, markPolicyAgnosticRL2024}. Finally, a set of work has looked at residual RL, where a base behavior cloned policy is kept frozen during training, and RL learns a residual policy on top to apply corrections to the base policy~\cite{ankileImitationRefinementResidual2024a, ankileResidualOffPolicyRL2025}. This work has been able to leverage powerful behavior cloning architectures like diffusion policies~\citep{chiDiffusionPolicyVisuomotor2023} for the frozen base policy, which other work has shown can be challenging to finetune directly~\citep{renDiffusionPolicyPolicy2024}. We draw insights from RLPD (\citet{ballEfficientOnlineReinforcement2023}), SERL (\citet{luoSERLSoftwareSuite2025}), ResiP (\citet{ankileImitationRefinementResidual2024a}) and ResFit (\citet{ankileResidualOffPolicyRL2025}) when designing our Off-Policy Residual RL framework.


\section{Methods}


In our problem setting, we are given a large set of human video demonstrations $\mathcal{V} = \{\langle o_0,o_1,\ldots,o_T\rangle\ldots\}$ and a small set of on-robot demonstrations $\mathcal{D} = \{\langle(s_0,a_0),(s_1,a_1),\ldots,(s_n,a_n)\rangle\ldots\}$ on a task $\mathcal{T}$ specified via a language tag. A policy $\pi$ trained only on $\mathcal{D}$ is not sufficiently performant on task $\mathcal{T}$, so we seek to derive a reward signal $R(s,a)$ such that if we adapted $\pi$ to maximize the value function $V(s_t) = E\Big[\sum_{t=0}^\infty \gamma^t R(s_t,\pi(s_t)\Big]$ computed with this reward, $\pi$ would increase its performance on task $\mathcal{T}$. This problem is particularly challenging as not only does the agent in $\mathcal{V}$ have a different action space from our robot, but we also cannot directly observe actions in $\mathcal{V}$, only the visual changes resulting from these actions. Note that we make a distinction between states ($s_t$) and observations ($o_t$), where states contain visual information from static cameras and robot-mounted cameras and proprioceptive information from the robot, while observations contain only a single camera observation. Observations can be extracted from both human videos $\mathcal{V}$ and robot data $\mathcal{D}$ while states are only available in robot data $\mathcal{D}$.

Our approach learns a motion prediction model $P_{t+1} = F_\theta(o_t,P_t,t/T)$ from $\mathcal{V}$ where $o_t$ is a visual observation at time $t$, $P_t$ is the locations of a set of points, and $t/T$ is a normalized task progress indicator. This model $F_\theta$ captures the ``preferences'' of the human demonstrators at each step in the task. For each attempt the robot makes at task $\mathcal{T}$, we measure the agreement between the predicted and observed motion between frames as our reward signal. Finally, an actor-critic algorithm uses this signal to estimate a value function and learn corrections to $\pi$ that increase its performance on task $\mathcal{T}$.

\subsection{Reward Learning as Modeling Expert Preferences}

The first stage of our approach is training our motion prediction model $F_\theta$ using human data $\mathcal{V}$. We first preprocess the data by tracking masks of task-relevant objects through each video, as well as tracking a grid of sampled points across the video. $F_\theta$ is then trained to output the location of a set of points at time $t+1$ given their locations at time $t$ as well as an observation from time $t$. To compute a reward signal for an episode of robot behavior, we extract observations $\langle o_0,o_1,\ldots,o_T\rangle$, detect and track masks for task-relevant objects, and track a set of points $\langle P^{track}_0, P^{track}_0, \ldots, P^{track}_T\rangle$ across the episode. Using these observations, point locations, and $F_\theta$, we compute a set of predicted point locations for each step in the episode $\langle P^{pred}_1, P^{pred}_2, \ldots, P^{pred}_T\rangle$. For each point $p \in P$, we compute a vector for how we predicted it to move between frames $t$ and $t+1$, $\Delta p^{pred}_t = p^{pred}_{t+1} - p^{pred}_t$, and how it was observed to move between frames $\Delta p^{track}_t = p^{track}_{t+1} - p^{track}_t$. With this information, we can compute the reward signal for frame t as:
\begin{equation}
    r_t = 
    \hspace{-7mm} 
    \sum_{
    \begin{array}{c}\Delta p^{pred}_t \in \Delta P^{pred}_t,\\ 
    \Delta p^{track}_t \in \Delta P^{track}_t 
    \end{array}}
    \hspace{-7mm}
    max\Bigg(\frac{\Delta p^{pred}_t \cdot \Delta p^{track}_t}{\|\Delta p^{pred}_t\|\|\Delta p^{track}_t\|},0\Bigg) 
\end{equation}

where $\Delta P^{pred}_t$ and $\Delta P^{track}_t$ are the set of predicted and tracked point deltas, respectively, at time $t$. Essentially, this reward computation is measuring the alignment (positive cosine similarity) between predicted and tracked point deltas at each time step. If we consider the predicted point deltas to capture the one-step preferences of the human demonstrator, i.e., how objects tend to move at each step in the task, then this reward signal rewards the robot for taking actions that match these preferences. See section \ref{sup_PPT_Details} in the appendix for further details on this implementation.

This approach has several advantages over prior work that measures the value of a visual state as its temporal distance to the goal. Prior work defines a value function as $V(o_t) = \sum_{t}^T -1$ for each demonstration video, where observations with further temporal distance from the goal have lower value and learn a model $V_\phi(o_t)$ to predict this value function conditioned on $o_t$. As mentioned, hesitations or non-time-optimal actions by the human demonstrator at time $t$ will negatively bias the value of all prior observations $\langle o_0, o_1, \ldots, o_{t-1}\rangle$. By measuring reward $r_t$ conditioned only on prediction and track deltas $\Delta P^{pred}_t$ and $\Delta P^{track}_t$ at time $t$, only the reward estimate at time $t$ is biased. $V_\phi(o_t)$, which is trained on only human video data, must also transfer its value estimates to videos of robots performing tasks, a significant distribution shift that may incur unexpected biases in predicted value. By only computing reward using point deltas on task-relevant objects, we lessen the distribution shift in jumping the embodiment gap from humans to robots and focus on the features that are consistent between these two domains. In addition, computing reward only across task-relevant objects incentivizes the robot to match the outcomes of human actions, not the actions themselves. This quality helps distinguish unsuccessful attempts that have approximately the right motions from genuinely successful attempts.

While our method predicts single step reward, instead of estimating long-term value like prior work, practically, this is a non-issue. Any modern reinforcement learning framework will first learn a value function as $Q_\psi(s_t,a_t) = R(s_t,a_t) + \gamma Q_\psi(s_{t+1},\pi(s_{t+1}))$ before performing any policy gradient or actor loss calculation, and this value function is likely to be more accurate as it is computed on the robot state distribution, and not the human video observation distribution. 

Finally, readers familiar with the no-regret learning literature may observe similarities between our approach and DAgger \cite{rossReductionImitationLearning2011}. Given an expert policy, DAgger trains a learner policy to match the expert's action in every state the learner visits. \citet{rossReductionImitationLearning2011} observe that a learner that matches the expert's preferences on the learner's state distribution is more likely to converge to a good policy than a learner who trains to match the expert's preferences on the expert's state distribution, as in typical behavior cloning.  A robot that optimizes a policy to maximize the reward function computed with our approach is learning to match the estimated expert preferences in its own state distribution.

\subsection{Point Prediction Model}

In this work, we use visual point tracks in a video as a representation of preferences for robot learning. Point tracking models take in a set of frames $\langle o_0,o_1,\ldots,o_T\rangle$ where $o_t \in \mathbb{R}^{H\times W\times 3}$ and a set of query points $\mathcal{P} = [p_1, p_2, \ldots, p_k]$ with $p_i = (x_t,y_t) \in \mathbb{R}^2$ where $(x_t,y_t)$ is the query point location at frame $t$. These models then output the image location of each point $p_i$ across the entire video $(x_t,y_t)_{t=1}^T$. Recent advances in point tracking models, such as CoTracker3~\citep{karaevCoTrackerItBetter2024, karaevCoTracker3SimplerBetter2024}, show remarkably robust performance across many ``in-the-wild" videos, leading us to choose this representation for our work. 

Our motion prediction model $F_\theta(o_t, P_t,\tau) = P_{t+1}$ parameterized by weights $\theta$ takes in an observation $o_t$ a set of query points $P_t$ and a normalized task time scalar $\tau = t/T$ and outputs the locations of the query points in the next frame $P_{t+1}$. For a video $\langle o_0,o_1,\ldots,o_T\rangle$ with tracked points $\langle P^{track}_0, P^{track}_0, \ldots, P^{track}_T \rangle$ we train $F_\theta$ with a simple regression loss:
\begin{equation}
    l(\theta) = \|F_\theta(o_t,P_t,t/T) - P_{t+1}\|^2
\end{equation}

We model $F_\theta$ with a modified version of the DiT architecture from the Diffusion Transformer paper \citep{peeblesScalableDiffusionModels2023}. While we train our model with regression, not diffusion, we find the DiT architecture is useful to model how a set of input points changes given various conditioning features.
While the standard DiT architecture conditions the main body of the transformer with feature-wise linear modification (FiLM), and previous work by \citet{bharadhwajTrack2ActPredictingPoint2024a} that predicted point motion with this architecture used a ResNet encoder with a single vector output to condition the model, we add a cross-attention layer with visual tokens from a frozen DinoV2 \citep{oquabDINOv2LearningRobust2024} pretrained vision transformer. We find that DinoV2 provides robust visual features that reduce the learning burden on our model, and the spatial information from visual tokens improves the prediction of point motion. This training is visualized in the yellow portion of figure \ref{fig:Intro}. 
Further training details can be found in Appendix section \ref{sup_PPT_Details}.

\subsection{Egocentric Human Video Preprocessing}

To learn from noisy and unstructured human video, our pipeline includes important preprocessing steps to extract the relevant inputs to our point prediction model $F_\theta$ from the human video dataset $\mathcal{V}$. For a task $\mathcal{T}$ specified by a language tag, we first retrieve a set of videos with labels matching $\mathcal{T}$ from the Ego4D \cite{graumanEgo4DWorld30002022a} and Epic Kitchens \citep{damenEPICKITCHENSDatasetCollection2020} datasets. While in this work, we retrieve based on matching language tags, we also saw success with using embeddings computed by the LaViLa model \citep{zhaoLearningVideoRepresentations2022a}, which would enable this work to scale to unlabeled videos. We use a combination of an open-vocabulary object detection model,  OWLViT2~\citep{mindererScalingOpenVocabularyObject2024}, a segmentation model, SAM2~\citep{raviSAM2Segment2024}, and a mask tracker, Cutie~\citep{chengPuttingObjectBack2024} to detect, segment, and track task-relevant objects. As the clip start and end times in large online video datasets are not always completely accurate, we trim portions of the video where task-relevant objects are not visible. We then track a grid of points through the video and, using the tracked object masks, separate these points into a set of object points and background points. To compensate for the significant camera motion in egocentric video, we subtract the mean of the background point motion between frames from both the object point deltas and the background point deltas. Finally, we train the model with a weighted mixture of object and background points. Further details on data preprocessing are available in section \ref{sup_PPT_Details}.


\subsection{Residual Reinforcement Learning}

To optimize a policy $\pi$ to maximize our reward function, we leverage methods in the learning from demonstration and practice literature. To create a learning pipeline that is both sample-efficient and stable, we combine insights from several prior works. Work by \citet{ankileImitationRefinementResidual2024a} trains a diffusion policy\citep{renDiffusionPolicyPolicy2024} on a set of demonstration data and then learns a residual policy with reinforcement learning that applies corrective actions $\Delta a$ to improve the performance of the base policy. Importantly, the base policy is frozen during this training, and only the residual corrective action is learned. While the original work used PPO, we found in experiments that if the residual policy ever pushed the robot too far outside the distribution of the demonstration data $\mathcal{D}$, the policy learning would usually collapse. To remedy this, we draw insights from work by \citep{ballEfficientOnlineReinforcement2023}, which identifies a set of design decisions to enable off-policy RL algorithms such as Soft Actor-Critic (SAC) \citep{guptaSoftActorCritic} to effectively leverage an offline dataset $\mathcal{D}$ along with an online buffer $\mathcal{B}$ of data collected from the robot interacting with the environment during training. Specifically, they take the minimum value across multiple critic networks and use LayerNorm in these critic networks to reduce the likelihood that the critic function overestimates value in low-data regions of task space. This decision reduces the likelihood of the actor falsely exploiting value overestimates. They also sample equally from $\mathcal{D}$ and $\mathcal{B}$ during training. We implement these two modifications, as well as pretraining the critic on offline data $\mathcal{D}$ labeled with rewards from our reward function to warm-start online finetuning. Importantly, for a given state-action pair $(s_t,a_t)$ in the offline data $\mathcal{D}$, we replace $a_t$ with $\Delta a_t$ by subtracting the output of the base policy from $a_t$. Our combined policy executed during online training is given by $\pi(s) = \pi_{base}(s) + \alpha \pi_{RL}(s)$ where $\pi_{base}$ is the frozen base policy learned with behavior cloning, $\pi_{RL}$ is the residual reinforcement policy, and alpha is a scalar constant that limits the magnitude of the residual actions. We find that even when using a base policy, training with the offline dataset $\mathcal{D}$ serves as a regularizer to prevent the actor from straying too far away from the demonstration data (as the residual actions for states in the demonstration data are close to 0). While freezing the base policy and applying a regularizer to ensure the learned policy does not deviate significantly from the data distribution of the base policy can potentially limit performance, we find that the stability improvements are well worth it, especially in a real-world finetuning scenario. Some of these design decisions also appear in work by \citet{luoSERLSoftwareSuite2025} and concurrent work by \citet{ankileResidualOffPolicyRL2025}. Overall, our experimentation here reaffirms what many others working in RL have found; regularization, both in learned value functions to prevent overestimation and in polices to prevent the policy from straying too far from regions of good data, is crucial to building stable and efficient RL pipelines. This training is visualized in the rightmost portion of figure \ref{fig:Intro}. Further details about our residual RL implementation can be found in Appendix section \ref{sup_RL_Details}.

\section{Results}
\subsection{Simulation Results}
\begin{figure*}[t!] 
    \centering
  \subfloat{%
       \includegraphics[width=0.49\linewidth]{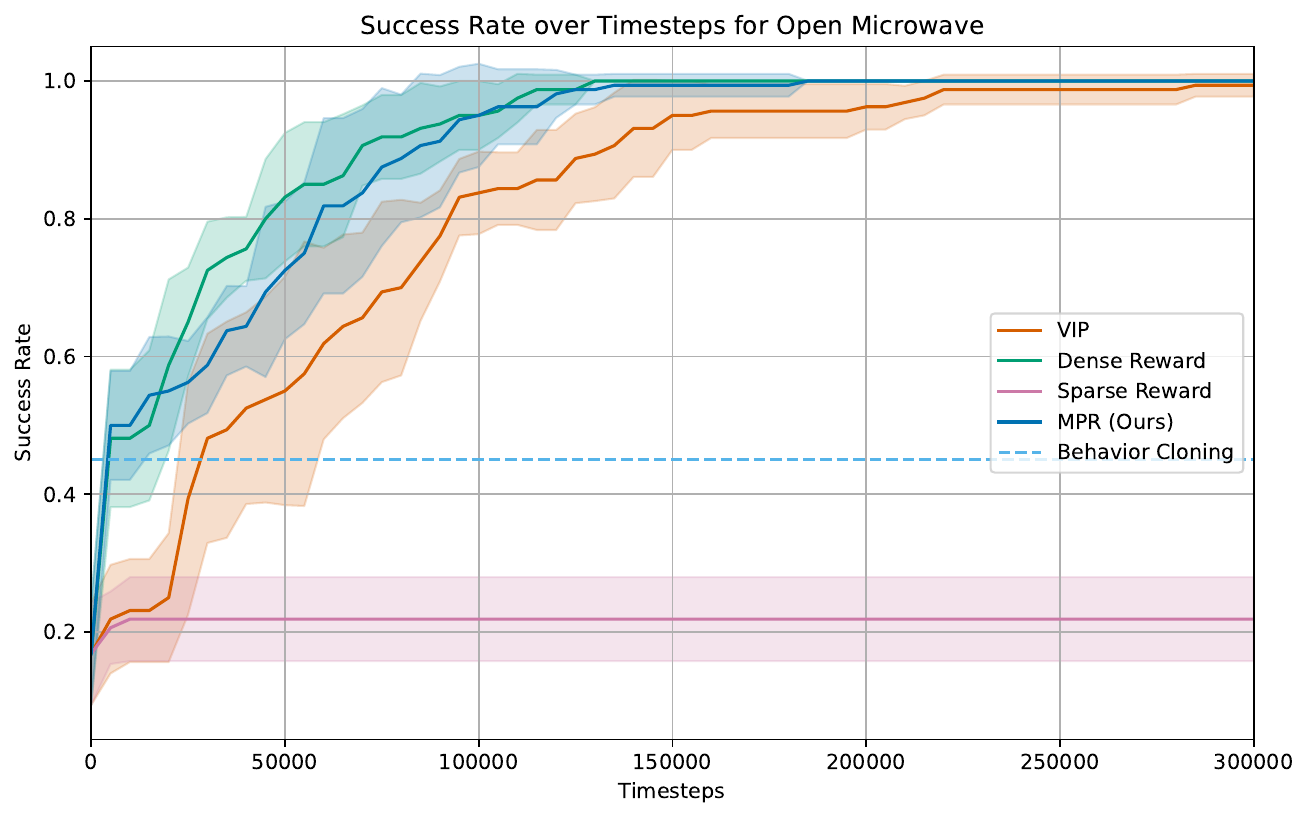}}
  \subfloat{%
        \includegraphics[width=0.49\linewidth]{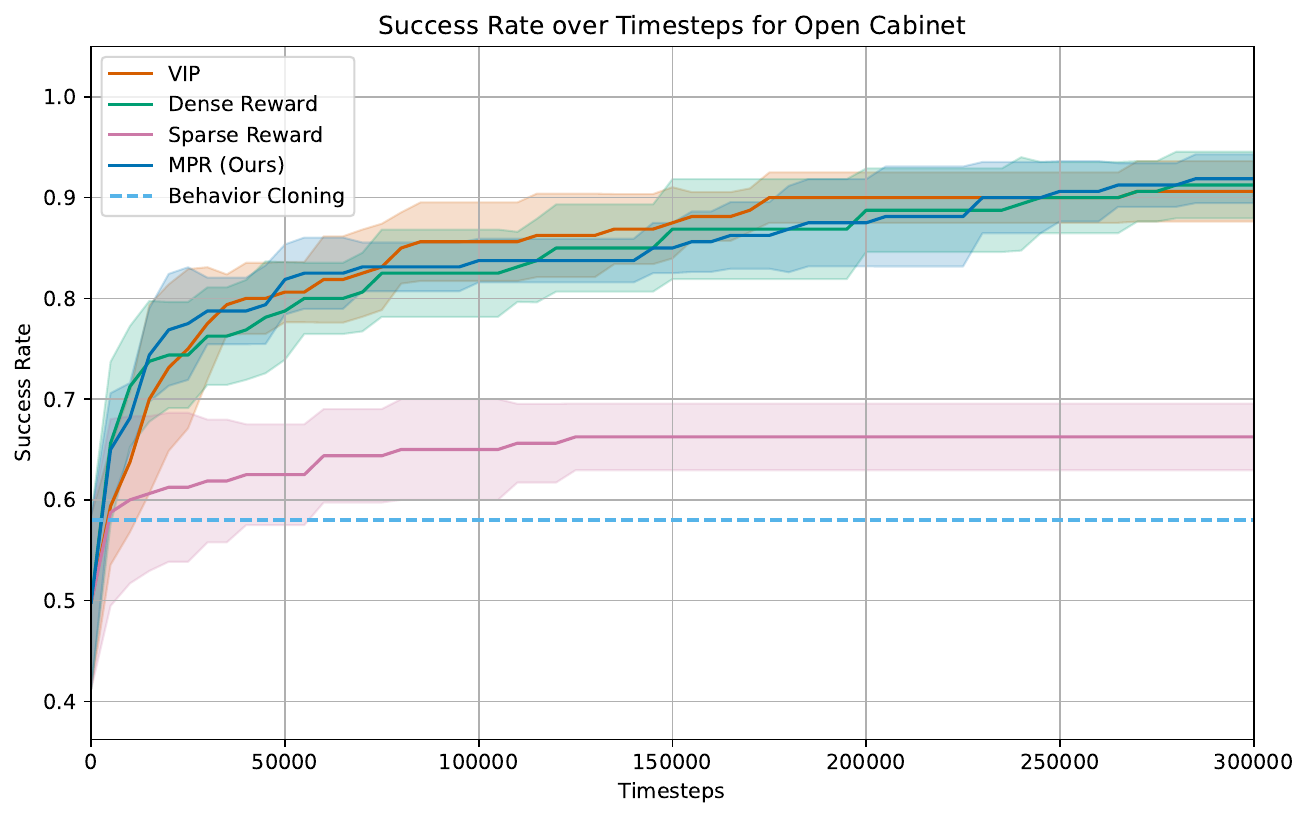}}

  \caption{\textbf{Success rates for various reward models across training timesteps in simulated tasks from the Franka Kitchen benchmark.} We evaluate different reward signals in our residual RL framework across two differnt simulated tasks. The reward signals include a sparse reward signal of 1 for task success and 0 otherwise, a handcrafted dense reward signal that measures progress to the goal using privileged simulation information, Value Implicit Pretraining (VIP)~\citep{maVIPUniversalVisual2023}, which is representative of the temporal distance class of reward learning methods, and our work, Motion Prediction Reward (MPR). The success rates shown are calculated across 20 evaluations per checkpoint and 8 different seeds for each method. Standard deviations across the 8 seeds are shaded, and the dashed line shows the base policy performance. While both MPR and VIP match the handcrafted sparse reward in the cabinet task, MPR outperforms VIP in the microwave tasks and closely tracks the handcrafted reward performance.}
  \label{fig:Sim_Result} 
  \vspace{-3mm}
\end{figure*}
\begin{figure}[t] 
    \centering
    \includegraphics[width=0.99\linewidth, trim={0cm 0cm 0cm 0cm},clip]{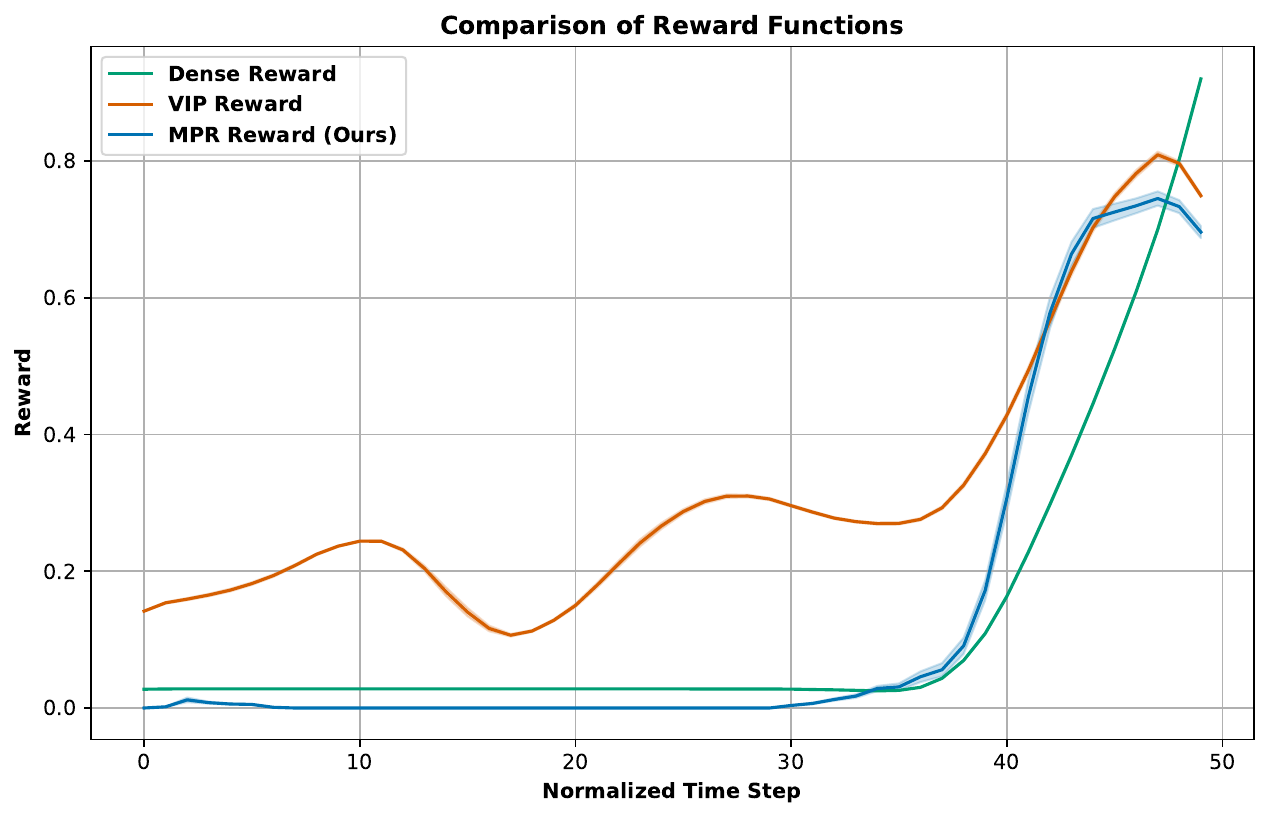}
  \caption{\textbf{Comparison of reward signals across 50 successful demonstrations for the simulated ``Open Microwave" task.} This plot shows the average estimated reward signal for our method (MPR), VIP~\cite{maVIPUniversalVisual2023}, and a handcrafted dense reward signal that uses privileged simulation information. Our reward signal closely tracks the handcrafted reward and shows very little bias or false positive results.}
  \label{fig:Sim_Rewards} 
  \vspace{-3mm}
\end{figure}

We first evaluate the performance of our approach on two tasks, ``Open Microwave" and ``Open Cabinet", in the simulated Franka Kitchen~\cite{guptaRelayPolicyLearning2019} benchmark. We chose these two tasks as they are well represented in the Ego4D~\citep{graumanEgo4DWorld30002022a} and Epic Kitchens~\citep{damenEPICKITCHENSDatasetCollection2020} egocentric video datasets from where we retrieve human video demonstrations. We compare our approach to several baselines. The first is a sparse reward function, which gives 1 on task success and 0 otherwise. Second, we compare with a hand-designed dense reward function that uses privileged state information (the positions and angles of relevant objects) in its reward computation. Finally, we compare with Value Implicit Pretraining (VIP)~\citep{maVIPUniversalVisual2023}, a representative approach from the class of temporal-distance-based value functions learning methods that we select for its easily available models and similar training data. We refer to our approach as Motion Prediction Reward (MPR) for experiments.

Each model uses a frozen base policy trained on 20 demonstrations from the simulated benchmark, and these demonstrations are also added to the offline data buffer $\mathcal{D}$. The models are trained for 300,000 simulated steps and evaluated every 50,000 steps for 20 episodes. The values plotted are the mean and standard deviations (shaded portions) of success rates across 8 seeds for each method. We plot the performance of the best checkpoint so far at each evaluation step. Note the methods initially underperform the base policy due to the randomly initialized residual policy which adds noise to the base policy actions. All training parameters can be found in the Appendix in section \ref{sup_BC_RL_Param}

As shown in Figure \ref{fig:Sim_Result}, while MPR, VIP, and the handcrafted dense reward function all perform similarly on the ``Open Cabinet" task, MPR clearly outperforms VIP in the ``Open Microwave" task and closely tracks the performance of the handcrafted dense reward function. Examining the estimated reward signals across 50 successful demos for the ``Open Microwave" in Figure \ref{fig:Sim_Rewards} task illustrates why. While our reward signal very closely tracks the handcrafted reward signal, VIP shows biases in the reward signal for certain states, which lead a learning agent into local minima and slow learning.

\subsection{Hardware Experiments}

\begin{figure}[t] 
    \centering
    \includegraphics[width=0.99\linewidth, trim={0cm 0cm 0cm 0cm},clip]{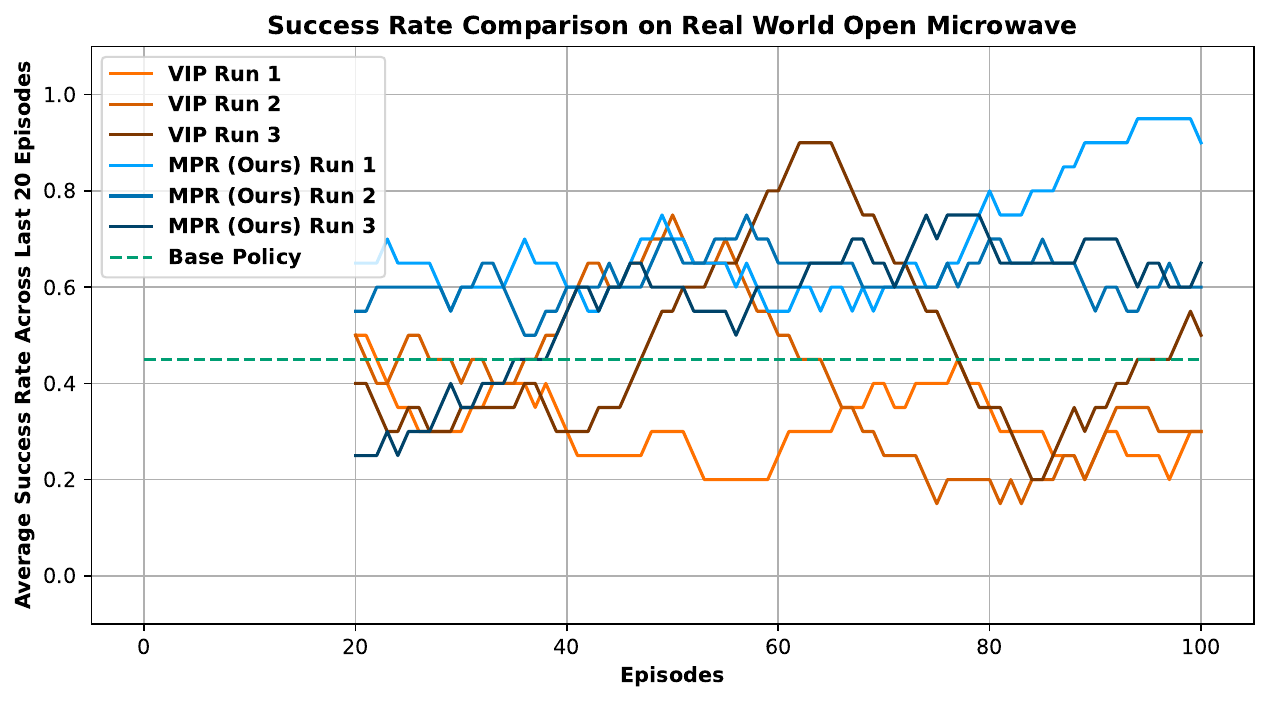}
  \caption{\textbf{Real-world training performance for the ``Open Microwave'' task across three runs each for VIP \citep{maVIPUniversalVisual2023} and our approach, Motion Prediction Reward (MPR).} All policies were initialized with the same base policy capable of completing the task 45\% of the time (9/20 attempts) and trained for 100 episodes (about an hour of wall clock time). Our residual RL framework leveraged the demo data used to train the base policy (10 demonstrations) and data collected online, both labeled with reward signals generated by each method. The plot shows a running average success rate across the last 20 episodes, with VIP in shades of orange and MPR in shades of blue. All MPR runs improve over the base policy and finish with an average success rate of 76.7\% across 20 evaluations on the final checkpoint. In contrast, the VIP runs face significant issues with stability and demonstrate ``unlearning'' behavior, finishing with an average success rate of 23.3\%, well below the base policy's performance.}
  \label{fig:Microwave_Training} 
  \vspace{-3mm}
\end{figure}
\begin{figure*}[t] 
    \centering
  \subfloat{%
       \includegraphics[width=0.49\linewidth]{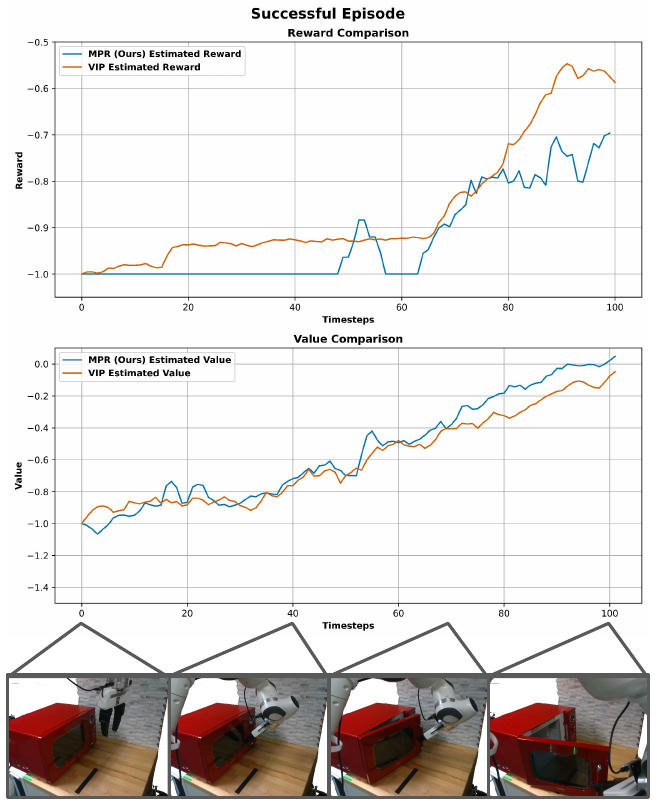}\label{fig:RV_Success}}
  \subfloat{%
        \includegraphics[width=0.49\linewidth]{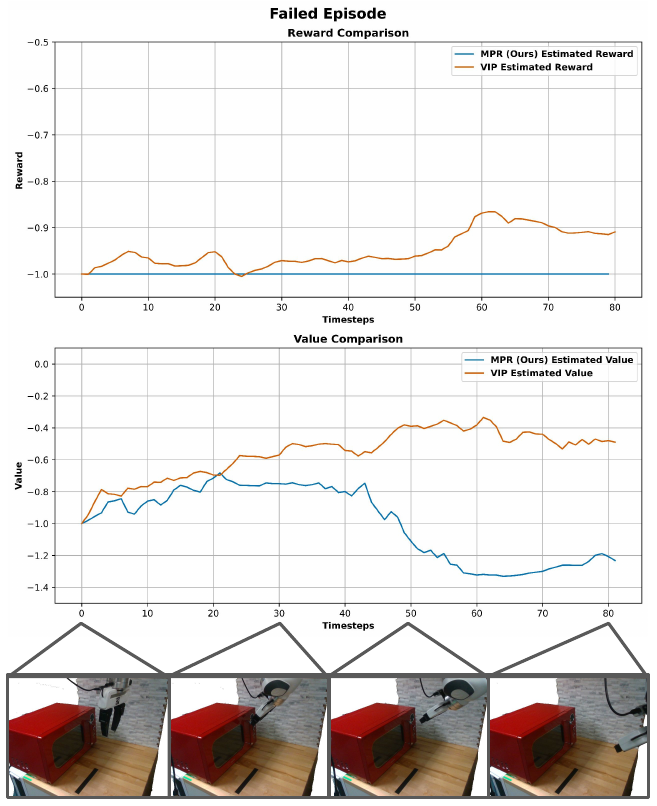}\label{fig:RV_Fail}}

  \caption{\textbf{Estimated reward signals and value estimates after training from VIP and MPR show VIP has trouble identifying failed episodes.} Estimated rewards (top) and values computed by learned value functions (bottom) after 100 episodes of training by each reward model for a successful (left) and failed (right) episode. Note value estimates are computed using the full robot state (world image, wrist image, end-effector pose, and base action) and the policy's action. Both approaches are capable of identifying a successful episode as shown in their reward and value estimates. However, VIP falsely assigns a moderate value to later actions and states in the failed episode, after the robot has missed grabbing the microwave handle, while MPR assigns no reward and a low value to the failed states. 
  }
  \label{fig:Reward_and_Value} 
  \vspace{-3mm}
\end{figure*}
To validate our reward learning approach on real robot hardware, we test it across three tasks. In ``Open Microwave", the robot must grasp the handle of a microwave to pull the door open. In ``Fold Cloth", the robot must fold a square cloth corner-to-corner into a triangle. And finally, in ``Wipe Counter," a robot must pick up a folded towel and use it to wipe coffee grounds off the counter into a bin. We use a Franka Research 3 equipped with UMI~\citep{chiUniversalManipulationInterface2024a} fingers and two Realsense D435 cameras, one looking over the robot's shoulder, called the world camera, and one mounted just above the robot's gripper, called the wrist camera. All observations for reward estimation are taken from the world camera. We control the robot in the space of relative position deltas, where the action space is an 8-vector consisting of position, quaternion rotation, and gripper width. The observation space consists of the RGB images from the world camera and wrist cameras, the end-effector pose and gripper width, and the base policy action. We design a gymnasium\citep{towersGymnasiumStandardInterface2025}-like interface to parse actions from the robot policy and communicate observations back. For each task, we train the base diffusion policy with 10 teleoperated demonstrations. We then give the robot a budget of 100 episodes of environment interaction to improve using the given reward function with a critic and policy network updates after each episode. These 100 episodes take an hour to an hour and a half of wall clock time. We perform three runs for each reward function on each task, and the performance is evaluated on the final checkpoint across 20 episodes. Relative to the simulation experiments, we increase the update-to-data ratio to 4 (at the end of an episode, we perform four times as many gradient steps on the critic and policy networks as steps in the episode) to accelerate learning, and pretrain the critic network for 10,000 iterations using the demonstration data $\mathcal{D}$ labeled with reward from each reward function. We also use a frozen pretrained DinoV2 encoder to process the image components of the state to accelerate learning. Even with these modifications, reinforcement learning on real robot hardware is very challenging as methods have a limited number of episodes and must learn in a very sample-efficient manner. Any bias or noise in a reward singal can significantly slow learning and limit the improvement the policy makes in the 100-episode window. Further details on the robot hardware setup are available in the Appendix in section \ref{sup_hardware}.

\begin{table}[t]
\centering
\scriptsize
\resizebox{.99\columnwidth}{!}{%


\begin{tabular}{lccc}
\hline
\textbf{Method} & \textbf{Open Microwave} & \textbf{Fold Cloth} & \textbf{Wipe Counter} \\ \hline
MPR (Ours)      & \textbf{76.7 $\pm$ 20.8\%}                       & \textbf{73.3 $\pm$ 7.6\%}                   & \textbf{56.7 $\pm$ 12.6 \%}                   \\
VIP             & 23.3 $\pm$ 32.1\%                      & 8.3$\pm$14.4\%                   & 21.7 $\pm$ 10.4 \%                   \\ 
\midrule
Base Policy & 45.0\% & 45.0\% & 40.0\%\\

\hline
\end{tabular}
}
\caption{\textbf{Success rates for the three tasks on the robot hardware.} We evaluate the policies after 100 episodes of training in the real world. These success rates represent 20 evaluations of the final checkpoint across three different runs per method and task. The base policy performance is also shown to demonstrate the relative improvement.}
\vspace{-3mm}
\label{tab:hardware_results}
\end{table}

\textbf{Motion Prediction Reward outperforms VIP across all hardware tasks.} Table \ref{tab:hardware_results} shows the performance of the final checkpoints for each method across the three tasks, and Figure \ref{fig:Microwave_Training} shows the performance of each run for the ``Open Microwave'' task. Plots of training performance for the other two tasks, as well as further experiments, can be found in section \ref{sup_additional_experiments} in the Appendix. Across all three tasks, we observe that policies trained with our motion prediction reward outperform those trained with VIP. Compared to 300,000 training steps in simulation, on hardware, 100 episodes translate to about 15,000 steps. This greatly reduced training budget highlights that MPR reward signals lead to more sample-efficient learning than those estimated by VIP. In our hardware experiments, we observed certain cases where false positives from VIP led training to collapse. In the ``fold cloth'' task, the robot must reach for one corner of the cloth, grasp it, lift it, and bring it over to the other corner, and place it down to complete the fold. VIP rewards measure distances between the current and goal states. In this task the goal state was the robot with its end effector above the two stacked corners of the cloth at the end of the task. In one run, the VIP rewards led the robot to skip grabbing the first corner of the cloth, and move its end effector directly to the opposite corner and stop there. This indicates that the VIP reward function failed to capture the salient feature in this task, the folded cloth, and instead fixated on the robot's position.

\textbf{Motion Prediction Reward enables a robot to improve its own performance in real hardware trials.}
In the ``Open Microwave'' and ``Fold Cloth'' tasks, we observe an increase in the success rate by 31.7 and 28.3 percentage points, respectively, over the base policies. In these tasks, with about an hour of training, MPR enables a robot to significantly increase its success rate while VIP is unable to match the performance of the base policy. In the ``Wipe Counter'' task, training with MPR rewards only yielded a 16.7\% improvement over the base policy. During training on this task, we observed that the base policy trained with only 10 demos was very sensitive to the reset conditions, and its performance was inconsistent. This inconsistency also led to residual RL training with both reward models being particularly unstable on this task, highlighting the importance of the base policy in residual RL.

\textbf{MPR reward estimation better discriminates between successful and failed episodes.} Figure \ref{fig:Microwave_Training} shows the average success rates in a 20-episode sliding window for each of the three runs with MPR Reward (shades of blue) and VIP Reward (shades of orange) on the ``Open Microwave". Examining this plot, we see that while VIP has spans where its performance increases, its performance decreases just as quickly after, indicating the value function and policy learned with VIP rewards are demonstrating forgetting behavior. To understand why this happens to policies trained with VIP rewards, but not policies with MPR rewards, we plotted both reward functions and the learned value functions at the end of real-world training across a successful and a failed episode in Figure \ref{fig:Reward_and_Value}. As shown in Figure \ref{fig:RV_Success}, both reward models can recognize a successful episode, and the learned value functions reflect this quality. In the failed episode, the robot misses the handle of the microwave but still pulls back. While the robot goes through the motions of the task, the desired result, opening the microwave, is not actually achieved. MPR reward estimation does not give this episode any reward, as the task-relevant object, the microwave, never moved, and the value function learned by MPR clearly identifies that the robot has missed the handle and assigns a very low value to subsequent states. However, VIP reward estimation still assigns this failed episode some positive reward, and the learned value function does not recognize that the task has failed, or identify the point of failure. This can lead the learned value function to falsely assign positive value to failed episodes, and lead an actor attempting to maximize the value function to a local maxima, where the robot goes through the motions of the task and gets some reward, but doesn't actually complete the task. We believe this property is the likely cause of the forgetting effect observed in training runs. As shown in the simulation results, more data mitigates this effect and enables the actor to find a more global maximia, but this makes policies learned with VIP reward less sample efficient, and reduces their efficacy in real-world training. 

This result also indicates that MPR is likely capturing more salient features of the reward estimation process than VIP does. This is a benefit of MPR focusing on task-relevant objects, and a drawback of VIP having to transfer the estimated value function across a large distribution shift from human to robot videos. In addition, human video demonstrations primarily contain successful demonstrations. For this reason, methods that measure value as temporal distance may have a harder time distinguishing genuinely successful episodes from episodes where the robot goes through the motions of the task but does not achieve success.


%
\section{Discussion}
\label{sec:discussion}
Our experiments show MPR significantly outperforms VIP across all tasks on real-hardware. However, VIP is a multitask policy conditioned via goal image, while MPR is trained for a specific task. In future work, we would like to explore a language-conditioned multi-task variant of MPR. Given the availability of human video data, and scaling in egocentric dataset size \citep{buildaiegocentric10k2025}, we believe this work represents progress towards a dense reward foundation model for robot learning.

We explored a very low data scenario in this work, with only 10 demonstrations and an hour of policy refinement with the given reward function. While we saw promising results in our experiments, such limited data can be a challenge for reinforcement learning algorithms. Given the recent progress in video models and world models for robotics, we would like to explore search in a world model as another method for optimizing a policy to maximize our learned reward.

Finally, there are several ways we could continue to improve the quality of MPR models. We could finetune these models on video from robot demonstrations, or collect in-domain human videos to adapt the motion prediction to the specific task. In developing this method, we also tested training our motion prediction models with diffusion rather than regression. While we saw more accurate motion prediction with the diffusion models, their inference time was significantly longer, and we felt the performance improvement didn't justify the slower inference. We would like to revisit these models and experiment with ways to accelerate inference, potentially leveraging new flow-matching architectures.

\section{Conclusion} 
\label{sec:conclusion}

In this work, we demonstrated an approach to learning reward signals from human video by modeling per-step human preferences as the predicted motion of object points. This approach enabled us to transfer information from human video demonstrations of tasks to robots attempting to learn the same task. We demonstrated that with just 10 demonstrations and an hour of real-world training, our reward model enabled a robot to autonomously increase its performance on a task. We also observed that our approach to reward learning from human video significantly outperformed prior work that learned temporal-distance based value functions. This work represents a promising new approach to reward learning from human video data, and we are excited to pursue it further.




\section*{Acknowledgments}


\bibliographystyle{plainnat}
\bibliography{references}

\begin{thebibliography}{57}
\providecommand{\natexlab}[1]{#1}
\providecommand{\url}[1]{\texttt{#1}}
\expandafter\ifx\csname urlstyle\endcsname\relax
  \providecommand{\doi}[1]{doi: #1}\else
  \providecommand{\doi}{doi: \begingroup \urlstyle{rm}\Url}\fi

\bibitem[AI(2025)]{buildaiegocentric10k2025}
Build AI.
\newblock Egocentric-10k, 2025.
\newblock URL \url{https://huggingface.co/datasets/builddotai/Egocentric-10K}.

\bibitem[Ankile et~al.(2024)Ankile, Simeonov, Shenfeld, Torne, and Agrawal]{ankileImitationRefinementResidual2024a}
Lars Ankile, Anthony Simeonov, Idan Shenfeld, Marcel Torne, and Pulkit Agrawal.
\newblock From {{Imitation}} to {{Refinement}} -- {{Residual RL}} for {{Precise Assembly}}, December 2024.

\bibitem[Ankile et~al.(2025)Ankile, Jiang, Duan, Shi, Abbeel, and Nagabandi]{ankileResidualOffPolicyRL2025}
Lars Ankile, Zhenyu Jiang, Rocky Duan, Guanya Shi, Pieter Abbeel, and Anusha Nagabandi.
\newblock Residual {{Off-Policy RL}} for {{Finetuning Behavior Cloning Policies}}, September 2025.

\bibitem[Assran et~al.(2025)Assran, Bardes, Fan, Garrido, Howes, Mojtaba, Komeili, Muckley, Rizvi, Roberts, Sinha, Zholus, Arnaud, Gejji, Martin, Hogan, Dugas, Bojanowski, Khalidov, Labatut, Massa, Szafraniec, Krishnakumar, Li, Ma, Chandar, Meier, LeCun, Rabbat, and Ballas]{assranVJEPA2SelfSupervised2025}
Mido Assran, Adrien Bardes, David Fan, Quentin Garrido, Russell Howes, Mojtaba, Komeili, Matthew Muckley, Ammar Rizvi, Claire Roberts, Koustuv Sinha, Artem Zholus, Sergio Arnaud, Abha Gejji, Ada Martin, Francois~Robert Hogan, Daniel Dugas, Piotr Bojanowski, Vasil Khalidov, Patrick Labatut, Francisco Massa, Marc Szafraniec, Kapil Krishnakumar, Yong Li, Xiaodong Ma, Sarath Chandar, Franziska Meier, Yann LeCun, Michael Rabbat, and Nicolas Ballas.
\newblock V-{{JEPA}} 2: {{Self-Supervised Video Models Enable Understanding}}, {{Prediction}} and {{Planning}}, June 2025.

\bibitem[Bahety et~al.(2024)Bahety, Mandikal, Abbatematteo, and {Mart{\'i}n-Mart{\'i}n}]{bahetyScrewMimicBimanualImitation2024a}
Arpit Bahety, Priyanka Mandikal, Ben Abbatematteo, and Roberto {Mart{\'i}n-Mart{\'i}n}.
\newblock {{ScrewMimic}}: {{Bimanual Imitation}} from {{Human Videos}} with {{Screw Space Projection}}, May 2024.

\bibitem[Ball et~al.(2023)Ball, Smith, Kostrikov, and Levine]{ballEfficientOnlineReinforcement2023}
Philip~J. Ball, Laura Smith, Ilya Kostrikov, and Sergey Levine.
\newblock Efficient {{Online Reinforcement Learning}} with {{Offline Data}}, May 2023.

\bibitem[Bentivegna et~al.()Bentivegna, Atkeson, and Cheng]{bentivegnaLearningObservationPractice}
Darrin~C Bentivegna, Christopher~G Atkeson, and Gordon Cheng.
\newblock Learning {{From Observation}} and {{Practice Using Primitives}}.

\bibitem[Bharadhwaj et~al.(2024)Bharadhwaj, Mottaghi, Gupta, and Tulsiani]{bharadhwajTrack2ActPredictingPoint2024a}
Homanga Bharadhwaj, Roozbeh Mottaghi, Abhinav Gupta, and Shubham Tulsiani.
\newblock {{Track2Act}}: {{Predicting Point Tracks}} from {{Internet Videos}} enables {{Generalizable Robot Manipulation}}, August 2024.

\bibitem[Cheng et~al.(2024)Cheng, Oh, Price, Lee, and Schwing]{chengPuttingObjectBack2024}
Ho~Kei Cheng, Seoung~Wug Oh, Brian Price, Joon-Young Lee, and Alexander Schwing.
\newblock Putting the {{Object Back}} into {{Video Object Segmentation}}, April 2024.

\bibitem[Chi et~al.(2023)Chi, Feng, Du, Xu, Cousineau, Burchfiel, and Song]{chiDiffusionPolicyVisuomotor2023}
Cheng Chi, Siyuan Feng, Yilun Du, Zhenjia Xu, Eric Cousineau, Benjamin Burchfiel, and Shuran Song.
\newblock Diffusion {{Policy}}: {{Visuomotor Policy Learning}} via {{Action Diffusion}}, June 2023.

\bibitem[Chi et~al.(2024)Chi, Xu, Pan, Cousineau, Burchfiel, Feng, Tedrake, and Song]{chiUniversalManipulationInterface2024a}
Cheng Chi, Zhenjia Xu, Chuer Pan, Eric Cousineau, Benjamin Burchfiel, Siyuan Feng, Russ Tedrake, and Shuran Song.
\newblock Universal {{Manipulation Interface}}: {{In-The-Wild Robot Teaching Without In-The-Wild Robots}}, March 2024.

\bibitem[Damen et~al.(2018)Damen, Doughty, Farinella, Fidler, Furnari, Kazakos, Moltisanti, Munro, Perrett, Price, and Wray]{damenScalingEgocentricVision2018a}
Dima Damen, Hazel Doughty, Giovanni~Maria Farinella, Sanja Fidler, Antonino Furnari, Evangelos Kazakos, Davide Moltisanti, Jonathan Munro, Toby Perrett, Will Price, and Michael Wray.
\newblock Scaling {{Egocentric Vision}}: {{The EPIC-KITCHENS Dataset}}, July 2018.

\bibitem[Damen et~al.(2020)Damen, Doughty, Farinella, Fidler, Furnari, Kazakos, Moltisanti, Munro, Perrett, Price, and Wray]{damenEPICKITCHENSDatasetCollection2020}
Dima Damen, Hazel Doughty, Giovanni~Maria Farinella, Sanja Fidler, Antonino Furnari, Evangelos Kazakos, Davide Moltisanti, Jonathan Munro, Toby Perrett, Will Price, and Michael Wray.
\newblock The {{EPIC-KITCHENS Dataset}}: {{Collection}}, {{Challenges}} and {{Baselines}}, April 2020.

\bibitem[Ghasemipour et~al.(2025)Ghasemipour, Wahid, Tompson, Sanketi, and Mordatch]{ghasemipourSelfImprovingEmbodiedFoundation2025}
Seyed Kamyar~Seyed Ghasemipour, Ayzaan Wahid, Jonathan Tompson, Pannag Sanketi, and Igor Mordatch.
\newblock Self-{{Improving Embodied Foundation Models}}, September 2025.

\bibitem[Grauman et~al.(2022)Grauman, Westbury, Byrne, Chavis, Furnari, Girdhar, Hamburger, Jiang, Liu, Liu, Martin, Nagarajan, Radosavovic, Ramakrishnan, Ryan, Sharma, Wray, Xu, Xu, Zhao, Bansal, Batra, Cartillier, Crane, Do, Doulaty, Erapalli, Feichtenhofer, Fragomeni, Fu, Gebreselasie, Gonzalez, Hillis, Huang, Huang, Jia, Khoo, Kolar, Kottur, Kumar, Landini, Li, Li, Li, Mangalam, Modhugu, Munro, Murrell, Nishiyasu, Price, Puentes, Ramazanova, Sari, Somasundaram, Southerland, Sugano, Tao, Vo, Wang, Wu, Yagi, Zhao, Zhu, Arbelaez, Crandall, Damen, Farinella, Fuegen, Ghanem, Ithapu, Jawahar, Joo, Kitani, Li, Newcombe, Oliva, Park, Rehg, Sato, Shi, Shou, Torralba, Torresani, Yan, and Malik]{graumanEgo4DWorld30002022a}
Kristen Grauman, Andrew Westbury, Eugene Byrne, Zachary Chavis, Antonino Furnari, Rohit Girdhar, Jackson Hamburger, Hao Jiang, Miao Liu, Xingyu Liu, Miguel Martin, Tushar Nagarajan, Ilija Radosavovic, Santhosh~Kumar Ramakrishnan, Fiona Ryan, Jayant Sharma, Michael Wray, Mengmeng Xu, Eric~Zhongcong Xu, Chen Zhao, Siddhant Bansal, Dhruv Batra, Vincent Cartillier, Sean Crane, Tien Do, Morrie Doulaty, Akshay Erapalli, Christoph Feichtenhofer, Adriano Fragomeni, Qichen Fu, Abrham Gebreselasie, Cristina Gonzalez, James Hillis, Xuhua Huang, Yifei Huang, Wenqi Jia, Weslie Khoo, Jachym Kolar, Satwik Kottur, Anurag Kumar, Federico Landini, Chao Li, Yanghao Li, Zhenqiang Li, Karttikeya Mangalam, Raghava Modhugu, Jonathan Munro, Tullie Murrell, Takumi Nishiyasu, Will Price, Paola~Ruiz Puentes, Merey Ramazanova, Leda Sari, Kiran Somasundaram, Audrey Southerland, Yusuke Sugano, Ruijie Tao, Minh Vo, Yuchen Wang, Xindi Wu, Takuma Yagi, Ziwei Zhao, Yunyi Zhu, Pablo Arbelaez, David Crandall, Dima Damen, Giovanni~Maria
  Farinella, Christian Fuegen, Bernard Ghanem, Vamsi~Krishna Ithapu, C.~V. Jawahar, Hanbyul Joo, Kris Kitani, Haizhou Li, Richard Newcombe, Aude Oliva, Hyun~Soo Park, James~M. Rehg, Yoichi Sato, Jianbo Shi, Mike~Zheng Shou, Antonio Torralba, Lorenzo Torresani, Mingfei Yan, and Jitendra Malik.
\newblock {{Ego4D}}: {{Around}} the {{World}} in 3,000 {{Hours}} of {{Egocentric Video}}, March 2022.

\bibitem[Gupta et~al.(2019)Gupta, Kumar, Lynch, Levine, and Hausman]{guptaRelayPolicyLearning2019}
Abhishek Gupta, Vikash Kumar, Corey Lynch, Sergey Levine, and Karol Hausman.
\newblock Relay {{Policy Learning}}: {{Solving Long-Horizon Tasks}} via {{Imitation}} and {{Reinforcement Learning}}, October 2019.

\bibitem[Gupta()]{guptaSoftActorCritic}
Ritwik Gupta.
\newblock Soft {{Actor Critic}}---{{Deep Reinforcement Learning}} with {{Real-World Robots}}.
\newblock http://bair.berkeley.edu/blog/2018/12/14/sac/.

\bibitem[Guzey et~al.(2024)Guzey, Dai, Savva, Bhirangi, and Pinto]{guzeyBridgingHumanRobot2024}
Irmak Guzey, Yinlong Dai, Georgy Savva, Raunaq Bhirangi, and Lerrel Pinto.
\newblock Bridging the {{Human}} to {{Robot Dexterity Gap}} through {{Object-Oriented Rewards}}, October 2024.

\bibitem[Heppert et~al.(2024)Heppert, Argus, Welschehold, Brox, and Valada]{heppertDITTODemonstrationImitation2024}
Nick Heppert, Max Argus, Tim Welschehold, Thomas Brox, and Abhinav Valada.
\newblock {{DITTO}}: {{Demonstration Imitation}} by {{Trajectory Transformation}}.
\newblock In \emph{2024 {{IEEE}}/{{RSJ International Conference}} on {{Intelligent Robots}} and {{Systems}} ({{IROS}})}, pages 7565--7572, October 2024.
\newblock \doi{10.1109/IROS58592.2024.10801982}.

\bibitem[Hu et~al.(2024)Hu, Mirchandani, and Sadigh]{huImitationBootstrappedReinforcement2024}
Hengyuan Hu, Suvir Mirchandani, and Dorsa Sadigh.
\newblock Imitation {{Bootstrapped Reinforcement Learning}}, May 2024.

\bibitem[Jain et~al.(2025)Jain, Mohta, Kim, Bhardwaj, Ren, Feng, Choudhury, and Swamy]{jainSmoothSeaNever2025}
Arnav~Kumar Jain, Vibhakar Mohta, Subin Kim, Atiksh Bhardwaj, Juntao Ren, Yunhai Feng, Sanjiban Choudhury, and Gokul Swamy.
\newblock A {{Smooth Sea Never Made}} a {{Skilled SAILOR}}: {{Robust Imitation}} via {{Learning}} to {{Search}}, October 2025.

\bibitem[Karaev et~al.(2024{\natexlab{a}})Karaev, Makarov, Wang, Neverova, Vedaldi, and Rupprecht]{karaevCoTracker3SimplerBetter2024}
Nikita Karaev, Iurii Makarov, Jianyuan Wang, Natalia Neverova, Andrea Vedaldi, and Christian Rupprecht.
\newblock {{CoTracker3}}: {{Simpler}} and {{Better Point Tracking}} by {{Pseudo-Labelling Real Videos}}, October 2024{\natexlab{a}}.

\bibitem[Karaev et~al.(2024{\natexlab{b}})Karaev, Rocco, Graham, Neverova, Vedaldi, and Rupprecht]{karaevCoTrackerItBetter2024}
Nikita Karaev, Ignacio Rocco, Benjamin Graham, Natalia Neverova, Andrea Vedaldi, and Christian Rupprecht.
\newblock {{CoTracker}}: {{It}} is {{Better}} to {{Track Together}}, October 2024{\natexlab{b}}.

\bibitem[Kareer et~al.(2024)Kareer, Patel, Punamiya, Mathur, Cheng, Wang, Hoffman, and Xu]{kareerEgoMimicScalingImitation2024}
Simar Kareer, Dhruv Patel, Ryan Punamiya, Pranay Mathur, Shuo Cheng, Chen Wang, Judy Hoffman, and Danfei Xu.
\newblock {{EgoMimic}}: {{Scaling Imitation Learning}} via {{Egocentric Video}}, October 2024.

\bibitem[Kim et~al.(2026)Kim, Gao, Lin, Lin, Ge, Lam, Liang, Song, Liu, Finn, and Gu]{kimCosmosPolicyFineTuning2026}
Moo~Jin Kim, Yihuai Gao, Tsung-Yi Lin, Yen-Chen Lin, Yunhao Ge, Grace Lam, Percy Liang, Shuran Song, Ming-Yu Liu, Chelsea Finn, and Jinwei Gu.
\newblock Cosmos {{Policy}}: {{Fine-Tuning Video Models}} for {{Visuomotor Control}} and {{Planning}}, January 2026.

\bibitem[Lee et~al.(2026)Lee, Wagenmaker, Pertsch, Liang, Levine, and Finn]{leeRoboRewardGeneralPurposeVisionLanguage2026}
Tony Lee, Andrew Wagenmaker, Karl Pertsch, Percy Liang, Sergey Levine, and Chelsea Finn.
\newblock {{RoboReward}}: {{General-Purpose Vision-Language Reward Models}} for {{Robotics}}, January 2026.

\bibitem[Lepert et~al.(2025)Lepert, Fang, and Bohg]{lepertMasqueradeLearningInthewild2025}
Marion Lepert, Jiaying Fang, and Jeannette Bohg.
\newblock Masquerade: {{Learning}} from {{In-the-wild Human Videos}} using {{Data-Editing}}, August 2025.

\bibitem[Li et~al.(2024)Li, Zhu, Xie, Jiang, Seo, Pavlakos, and Zhu]{liOKAMITeachingHumanoid2024}
Jinhan Li, Yifeng Zhu, Yuqi Xie, Zhenyu Jiang, Mingyo Seo, Georgios Pavlakos, and Yuke Zhu.
\newblock {{OKAMI}}: {{Teaching Humanoid Robots Manipulation Skills}} through {{Single Video Imitation}}, October 2024.

\bibitem[Liu et~al.(2025{\natexlab{a}})Liu, Adeniji, Zhan, Haldar, Bhirangi, Abbeel, and Pinto]{liuEgoZeroRobotLearning2025}
Vincent Liu, Ademi Adeniji, Haotian Zhan, Siddhant Haldar, Raunaq Bhirangi, Pieter Abbeel, and Lerrel Pinto.
\newblock {{EgoZero}}: {{Robot Learning}} from {{Smart Glasses}}, June 2025{\natexlab{a}}.

\bibitem[Liu et~al.(2025{\natexlab{b}})Liu, Wen, Hu, Jayaraman, and Gao]{liuTimeRewarderLearningDense2025}
Yuyang Liu, Chuan Wen, Yihang Hu, Dinesh Jayaraman, and Yang Gao.
\newblock {{TimeRewarder}}: {{Learning Dense Reward}} from {{Passive Videos}} via {{Frame-wise Temporal Distance}}, September 2025{\natexlab{b}}.

\bibitem[Luo et~al.(2025{\natexlab{a}})Luo, Hu, Xu, Tan, Berg, Sharma, Schaal, Finn, Gupta, and Levine]{luoSERLSoftwareSuite2025}
Jianlan Luo, Zheyuan Hu, Charles Xu, You~Liang Tan, Jacob Berg, Archit Sharma, Stefan Schaal, Chelsea Finn, Abhishek Gupta, and Sergey Levine.
\newblock {{SERL}}: {{A Software Suite}} for {{Sample-Efficient Robotic Reinforcement Learning}}, March 2025{\natexlab{a}}.

\bibitem[Luo et~al.(2025{\natexlab{b}})Luo, Xu, Wu, and Levine]{luoPreciseDexterousRobotic2025}
Jianlan Luo, Charles Xu, Jeffrey Wu, and Sergey Levine.
\newblock Precise and {{Dexterous Robotic Manipulation}} via {{Human-in-the-Loop Reinforcement Learning}}, March 2025{\natexlab{b}}.

\bibitem[Ma et~al.(2023{\natexlab{a}})Ma, Liang, Som, Kumar, Zhang, Bastani, and Jayaraman]{maLIVLanguageImageRepresentations2023b}
Yecheng~Jason Ma, William Liang, Vaidehi Som, Vikash Kumar, Amy Zhang, Osbert Bastani, and Dinesh Jayaraman.
\newblock {{LIV}}: {{Language-Image Representations}} and {{Rewards}} for {{Robotic Control}}, June 2023{\natexlab{a}}.

\bibitem[Ma et~al.(2023{\natexlab{b}})Ma, Sodhani, Jayaraman, Bastani, Kumar, and Zhang]{maVIPUniversalVisual2023}
Yecheng~Jason Ma, Shagun Sodhani, Dinesh Jayaraman, Osbert Bastani, Vikash Kumar, and Amy Zhang.
\newblock {{VIP}}: {{Towards Universal Visual Reward}} and {{Representation}} via {{Value-Implicit Pre-Training}}, March 2023{\natexlab{b}}.

\bibitem[Mahmoudieh et~al.(2022)Mahmoudieh, Pathak, and Darrell]{mahmoudiehZeroShotRewardSpecification2022a}
Parsa Mahmoudieh, Deepak Pathak, and Trevor Darrell.
\newblock Zero-{{Shot Reward Specification}} via {{Grounded Natural Language}}.
\newblock In \emph{Proceedings of the 39th {{International Conference}} on {{Machine Learning}}}, pages 14743--14752. PMLR, June 2022.

\bibitem[Majumdar et~al.(2024)Majumdar, Yadav, Arnaud, Ma, Chen, Silwal, Jain, Berges, Abbeel, Malik, Batra, Lin, Maksymets, Rajeswaran, and Meier]{majumdarWhereAreWe2024a}
Arjun Majumdar, Karmesh Yadav, Sergio Arnaud, Yecheng~Jason Ma, Claire Chen, Sneha Silwal, Aryan Jain, Vincent-Pierre Berges, Pieter Abbeel, Jitendra Malik, Dhruv Batra, Yixin Lin, Oleksandr Maksymets, Aravind Rajeswaran, and Franziska Meier.
\newblock Where are we in the search for an {{Artificial Visual Cortex}} for {{Embodied Intelligence}}?, February 2024.

\bibitem[Mandikal et~al.(2025)Mandikal, Hu, Dass, Majumder, {Mart{\'i}n-Mart{\'i}n}, and Grauman]{mandikalMashSpreadSlice2025}
Priyanka Mandikal, Jiaheng Hu, Shivin Dass, Sagnik Majumder, Roberto {Mart{\'i}n-Mart{\'i}n}, and Kristen Grauman.
\newblock Mash, {{Spread}}, {{Slice}}! {{Learning}} to {{Manipulate Object States}} via {{Visual Spatial Progress}}, September 2025.

\bibitem[Mark et~al.(2024)Mark, Gao, Sampaio, Srirama, Sharma, Finn, and Kumar]{markPolicyAgnosticRL2024}
Max~Sobol Mark, Tian Gao, Georgia~Gabriela Sampaio, Mohan~Kumar Srirama, Archit Sharma, Chelsea Finn, and Aviral Kumar.
\newblock Policy {{Agnostic RL}}: {{Offline RL}} and {{Online RL Fine-Tuning}} of {{Any Class}} and {{Backbone}}, December 2024.

\bibitem[Milikic et~al.(2025)Milikic, Patel, and Frey]{milikicVLDVisualLanguage2025}
Lazar Milikic, Manthan Patel, and Jonas Frey.
\newblock {{VLD}}: {{Visual Language Goal Distance}} for {{Reinforcement Learning Navigation}}, December 2025.

\bibitem[Minderer et~al.(2024)Minderer, Gritsenko, and Houlsby]{mindererScalingOpenVocabularyObject2024}
Matthias Minderer, Alexey Gritsenko, and Neil Houlsby.
\newblock Scaling {{Open-Vocabulary Object Detection}}, May 2024.

\bibitem[Nair et~al.(2022)Nair, Rajeswaran, Kumar, Finn, and Gupta]{nairR3MUniversalVisual2022d}
Suraj Nair, Aravind Rajeswaran, Vikash Kumar, Chelsea Finn, and Abhinav Gupta.
\newblock {{R3M}}: {{A Universal Visual Representation}} for {{Robot Manipulation}}, November 2022.

\bibitem[Oquab et~al.(2024)Oquab, Darcet, Moutakanni, Vo, Szafraniec, Khalidov, Fernandez, Haziza, Massa, {El-Nouby}, Assran, Ballas, Galuba, Howes, Huang, Li, Misra, Rabbat, Sharma, Synnaeve, Xu, Jegou, Mairal, Labatut, Joulin, and Bojanowski]{oquabDINOv2LearningRobust2024}
Maxime Oquab, Timoth{\'e}e Darcet, Th{\'e}o Moutakanni, Huy Vo, Marc Szafraniec, Vasil Khalidov, Pierre Fernandez, Daniel Haziza, Francisco Massa, Alaaeldin {El-Nouby}, Mahmoud Assran, Nicolas Ballas, Wojciech Galuba, Russell Howes, Po-Yao Huang, Shang-Wen Li, Ishan Misra, Michael Rabbat, Vasu Sharma, Gabriel Synnaeve, Hu~Xu, Herv{\'e} Jegou, Julien Mairal, Patrick Labatut, Armand Joulin, and Piotr Bojanowski.
\newblock {{DINOv2}}: {{Learning Robust Visual Features}} without {{Supervision}}, February 2024.

\bibitem[Palo and Johns(2024)]{paloDINOBotRobotManipulation2024}
Norman~Di Palo and Edward Johns.
\newblock {{DINOBot}}: {{Robot Manipulation}} via {{Retrieval}} and {{Alignment}} with {{Vision Foundation Models}}, February 2024.

\bibitem[Peebles and Xie(2023)]{peeblesScalableDiffusionModels2023}
William Peebles and Saining Xie.
\newblock Scalable {{Diffusion Models}} with {{Transformers}}, March 2023.

\bibitem[Qin et~al.(2022)Qin, Wu, Liu, Jiang, Yang, Fu, and Wang]{qinDexMVImitationLearning2022}
Yuzhe Qin, Yueh-Hua Wu, Shaowei Liu, Hanwen Jiang, Ruihan Yang, Yang Fu, and Xiaolong Wang.
\newblock {{DexMV}}: {{Imitation Learning}} for {{Dexterous Manipulation}} from {{Human Videos}}, July 2022.

\bibitem[Ravi et~al.(2024)Ravi, Gabeur, Hu, Hu, Ryali, Ma, Khedr, R{\"a}dle, Rolland, Gustafson, Mintun, Pan, Alwala, Carion, Wu, Girshick, Doll{\'a}r, and Feichtenhofer]{raviSAM2Segment2024}
Nikhila Ravi, Valentin Gabeur, Yuan-Ting Hu, Ronghang Hu, Chaitanya Ryali, Tengyu Ma, Haitham Khedr, Roman R{\"a}dle, Chloe Rolland, Laura Gustafson, Eric Mintun, Junting Pan, Kalyan~Vasudev Alwala, Nicolas Carion, Chao-Yuan Wu, Ross Girshick, Piotr Doll{\'a}r, and Christoph Feichtenhofer.
\newblock {{SAM}} 2: {{Segment Anything}} in {{Images}} and {{Videos}}, October 2024.

\bibitem[Ren et~al.(2024)Ren, Lidard, Ankile, Simeonov, Agrawal, Majumdar, Burchfiel, Dai, and Simchowitz]{renDiffusionPolicyPolicy2024}
Allen~Z. Ren, Justin Lidard, Lars~L. Ankile, Anthony Simeonov, Pulkit Agrawal, Anirudha Majumdar, Benjamin Burchfiel, Hongkai Dai, and Max Simchowitz.
\newblock Diffusion {{Policy Policy Optimization}}, December 2024.

\bibitem[Ren et~al.(2025)Ren, Sundaresan, Sadigh, Choudhury, and Bohg]{renMotionTracksUnified2025}
Juntao Ren, Priya Sundaresan, Dorsa Sadigh, Sanjiban Choudhury, and Jeannette Bohg.
\newblock Motion {{Tracks}}: {{A Unified Representation}} for {{Human-Robot Transfer}} in {{Few-Shot Imitation Learning}}, October 2025.

\bibitem[Ross et~al.(2011)Ross, Gordon, and Bagnell]{rossReductionImitationLearning2011}
Stephane Ross, Geoffrey~J. Gordon, and J.~Andrew Bagnell.
\newblock A {{Reduction}} of {{Imitation Learning}} and {{Structured Prediction}} to {{No-Regret Online Learning}}, March 2011.

\bibitem[Sermanet et~al.(2018)Sermanet, Lynch, Chebotar, Hsu, Jang, Schaal, and Levine]{sermanetTimeContrastiveNetworksSelfSupervised2018a}
Pierre Sermanet, Corey Lynch, Yevgen Chebotar, Jasmine Hsu, Eric Jang, Stefan Schaal, and Sergey Levine.
\newblock Time-{{Contrastive Networks}}: {{Self-Supervised Learning}} from {{Video}}, March 2018.

\bibitem[Shao et~al.()Shao, Migimatsu, Zhang, Yang, and Bohg]{shaoConcept2RobotLearningManipulation}
Lin Shao, Toki Migimatsu, Qiang Zhang, Karen Yang, and Jeannette Bohg.
\newblock {{Concept2Robot}}: {{Learning Manipulation Concepts}} from {{Instructions}} and {{Human Demonstrations}}.

\bibitem[Shen et~al.(2025)Shen, Wei, Du, Liang, Lu, Yang, Zheng, and Guo]{shenVideoVLAVideoGenerators2025}
Yichao Shen, Fangyun Wei, Zhiying Du, Yaobo Liang, Yan Lu, Jiaolong Yang, Nanning Zheng, and Baining Guo.
\newblock {{VideoVLA}}: {{Video Generators Can Be Generalizable Robot Manipulators}}, December 2025.

\bibitem[Towers et~al.(2025)Towers, Kwiatkowski, Terry, Balis, Cola, Deleu, Goul{\~a}o, Kallinteris, Krimmel, KG, {Perez-Vicente}, Pierr{\'e}, Schulhoff, Tai, Tan, and Younis]{towersGymnasiumStandardInterface2025}
Mark Towers, Ariel Kwiatkowski, Jordan Terry, John~U. Balis, Gianluca~De Cola, Tristan Deleu, Manuel Goul{\~a}o, Andreas Kallinteris, Markus Krimmel, Arjun KG, Rodrigo {Perez-Vicente}, Andrea Pierr{\'e}, Sander Schulhoff, Jun~Jet Tai, Hannah Tan, and Omar~G. Younis.
\newblock Gymnasium: {{A Standard Interface}} for {{Reinforcement Learning Environments}}, November 2025.

\bibitem[Wang et~al.(2025)Wang, Verghese, and Schneider]{wangLatentPolicySteering2025}
Yiqi Wang, Mrinal Verghese, and Jeff Schneider.
\newblock Latent {{Policy Steering}} with {{Embodiment-Agnostic Pretrained World Models}}, September 2025.

\bibitem[Yamaguchi et~al.(2014)Yamaguchi, Atkeson, Niekum, and Ogasawara]{yamaguchiLearningPouringSkills2014}
Akihiko Yamaguchi, Christopher~G. Atkeson, Scott Niekum, and Tsukasa Ogasawara.
\newblock Learning pouring skills from demonstration and practice.
\newblock In \emph{2014 {{IEEE-RAS International Conference}} on {{Humanoid Robots}}}, pages 908--915, Madrid, Spain, November 2014. IEEE.
\newblock ISBN 978-1-4799-7174-9.
\newblock \doi{10.1109/HUMANOIDS.2014.7041472}.

\bibitem[Yuan et~al.(2025)Yuan, Wei, Gu, Hua, Liang, Chen, and Xu]{yuanHERMESHumantoRobotEmbodied2025}
Zhecheng Yuan, Tianming Wei, Langzhe Gu, Pu~Hua, Tianhai Liang, Yuanpei Chen, and Huazhe Xu.
\newblock {{HERMES}}: {{Human-to-Robot Embodied Learning}} from {{Multi-Source Motion Data}} for {{Mobile Dexterous Manipulation}}, August 2025.

\bibitem[Zhao et~al.(2022)Zhao, Misra, Kr{\"a}henb{\"u}hl, and Girdhar]{zhaoLearningVideoRepresentations2022a}
Yue Zhao, Ishan Misra, Philipp Kr{\"a}henb{\"u}hl, and Rohit Girdhar.
\newblock Learning {{Video Representations}} from {{Large Language Models}}, December 2022.

\end{thebibliography}

\clearpage

\noindent \textbf{This appendix is organized as follows:} \\
\noindent 7. Motion Prediction Model Training Details

\noindent 8. Residual RL Details

\noindent 9. Hardware Details

\noindent 10. Behavior Cloning and RL Parameters

\noindent 11. Additional Experimental Data

\section{Motion Prediction Model Training Details}
\label{sup_PPT_Details}
Our motion prediction model training pipeline includes the following steps
\subsection{Task-Relevant Object Mask Tracking}
We use OwlViT2~\citep{mindererScalingOpenVocabularyObject2024} to get the highest likelihood bounding box for a desired object and then prompt SAM2~\citep{raviSAM2Segment2024} with this box to get an object mask. We run this detection for five frames equally sampled throughout the video and use the frame with the highest detection score. We then propagate this mask forward and backward in time using Cutie~\citep{chengPuttingObjectBack2024}, a mask tracker. Using the estimated mask, we find a square 256x256 crop of the video such that the task-relevant object is always visible. If needed, we crop a larger square and then downsample to 256x256 to ensure the task-relevant objects are visible throughout the whole video.
\subsection{Point Tracking}
We use CoTracker3~\citep{karaevCoTracker3SimplerBetter2024} to track an evenly sampled grid of 32x32 points in the video. CoTracker provides point locations and estimated visibility. We keep track of the proportion of visible points on the task-relevant objects, and if more than 30\% of them become no longer visible, we resample the grid of points. This helps account for cases where an object moves or the camera perspective shifts, causing us to lose track of points. We observed that CoTracker's performance decreases with very short horizon tracks, so we set a threshold of 30\% to balance good point density with minimal point resamples. We construct our training dataset where the inputs are the current and previous frames from the video, the point locations in the current frame, and the current frame number divided by the total length of the video, and the targets are the point locations in the next frame. We use an input of two frames to give the model context for any existing motion. We remove any frames from the training data where the max point motion is below 0.5 pixels, to avoid cases where frames are duplicated or there is no motion. We also normalize the point locations to be between zero and one.
\subsection{Input Preparation}
Using the masks of each frame and the point locations, we designate points in the mask as object points and points outside the mask as background points. We calculate the average vector of background point motion and subtract it from the object point motion. This helps compensate for the camera motion in egocentric video. At each frame, we use all object points and sample half the number of object points from the set of background points. This focuses learning on object point prediction, but still requires the network to compensate for any residual background motion. During initial testing, we found this mixture to be more effective than sampling all points or only object points. To make uniform batches during training, we repeat this set of points until we hit 300 points. We preprocess frames with the standard ImageNet normalization.
\subsection{Training parameters}
\begin{table}[h]
\begin{tabular}{lc}
\hline
Parameter     & Value                      \\ \hline
Epochs        & 1000                       \\
Batch Size    & 20                         \\
Learning Rate & $1e^{-4}$ \\
Val Ratio     & 0.1                        \\
Optimizer     & AdamW                      \\
Architecture  & DiT                         \\
\# Params     & 560M                       \\ \hline
\end{tabular}
\caption{\textbf{Motion Prediction Transformer training parameters.}}
\end{table}
\subsection{Inference}
We use the same input preparation process during inference, with the exception that we don't subtract background point motion, as our camera is fixed during inference. Our reward function computes a reward value between zero and one. During testing, we found that with a positive reward, the robot would become disinclined to complete the task, as completing the task ended the episode. Instead, it would keep manipulating task-relevant objects to try to get more positive rewards. To discourage this behavior, we just subtracted one from the reward output to shift the range to negative one to zero.
\subsection{Hardware}
All training was done on two RTX 4090s with 24GB of VRAM each. Training takes 12-48 hours, depending on the number of videos in the training set. Inference was done using one 4090, with the mask tracking stack and CoTracker running on the other 4090. Inference with full masking, tracking, and computation takes 5-10 seconds per episode.

\section{Residual RL Details}
\label{sup_RL_Details}
We implement our residual RL framework as a modified version of Soft Actor Critic (SAC)~\citep{guptaSoftActorCritic}. We primarily follow the modifications outlined in RLPD from~\citet{ballEfficientOnlineReinforcement2023} and apply them to the implementation of SAC in Stable Baselines 3 (SB3). SB3 already includes training dual critic networks and taking a minimum across critics for value estimation. Additionally, we add LayerNorm after all but the last layers of the critic network. We find that this has a significant impact on learning stability. Following RLPD, we sample evenly from the offline buffer containing the demonstration data and the online buffer containing collected transitions. Following \citet{ankileImitationRefinementResidual2024a}, we apply orthogonal initialization to actor weights with 0 bias. Full RL parameters can be found in Section \ref{sup_BC_RL_Param} of this appendix.

For efficiency and smooth execution on hardware, we perform reward computation and training at the end of each episode. We write a Gym wrapper to cache images from execution, and an SB3 callback to run after episode collection and compute reward signals.

\section{Hardware Details}
\label{sup_hardware}
We use a Franka Research 3 robot for all hardware experiments. This robot is equipped with two RealSense D435 cameras, one in a fixed ``over-the-shoulder'' location and one attached to the wrist joint. The robot is equipped with a parallel jaw gripper and has its provided fingers replaced with compliant UMI fingers~\citep{chiUniversalManipulationInterface2024a}. We use the Polymetis library to control our Franka robot. Polymetis leverages a separate control PC with a real-time kernel patch running a low-level controller at a higher frequency, and then an inference PC running a learned controller at a lower frequency. We run impedance control for both data collection and inference. This impedance controller runs at approximately 1000Hz on the control PC and provides high responsiveness while also preventing the robot from damaging itself or the environment. The impedance controller also lets the robot modulate force by adjusting its set point. We run our learned policy at approximately 10 Hz. During RL training and inference, we apply an additional damping factor of 0.1 on the magnitude of all commanded rotations. High-magnitude rotation commands can cause errors in the controller, and we found that rotation was particularly affected by action noise during training. This would add significant noise to the position of the end-effector and slow training. Additionally, total rotation commands are clipped to have a magnitude no greater than 0.25 radians.

\section{Behavior Cloning and RL Parameters}
\label{sup_BC_RL_Param}
\subsection{Behavior Cloning Parameters}
\begin{table}[h]
\begin{tabular}{lc}
\hline
Parameter                 & Value               \\ \hline
Epochs                    & 1000                \\
Batch Size                & 64                  \\
Learning Rate             & $1e^{-4}$             \\
Architecture              & Conditional UNet 1D \\
UNet Parameters           & 94.8M               \\ \hline
Optimizer                 & AdamW               \\
Optimizer Weight Decay    & $1e^{-6}$            \\
Optimizer Betas           & 0.9, 0.95           \\
LR Schedule               & Cosine              \\ \hline
Diffusion Steps           & 50                  \\
Diffusion Scheduler       & DDPM                \\
Diffusion Beta Start      & $1e^{-4}$             \\
Diffusion Beta End        & $2e^{-2}$             \\
Diffusion Beta Schedule   & Squared Cosine      \\ \hline
Vision Encoder            & ResNet18            \\
Vision Encoder Parameters & 11.7M               \\ \hline
EMA Model Power           & 0.75                \\ \hline
Observation Horizon       & 2                   \\
Prediction Horizon        & 16                  \\
Action Horizon            & 8                  \\\hline
\end{tabular}
\caption{\textbf{Diffusion Policy Behavior Cloning Parameters.}}
\end{table}
\subsection{RL Parameters}
Note, we only include parameters here that are different than their default values. To find a full list of SAC parameters, see the Stable Baselines 3 documentation.
\begin{table}[h]
\begin{tabular}{lc}
\hline
Parameter             & Value                   \\ \hline
Batch Size            & 64                      \\
Learning Rate         & $1e^{-4}$               \\
$\gamma$              & 0.99                    \\
Offline Ratio         & 0.5                     \\
Learning Starts       & 1024                    \\
Update to Data Ratio  & 4                       \\
Train Frequency       & Every Episode           \\
Network Hidden Size   & 256                     \\
Network Hidden Layers & 3                       \\
Vision Encoder        & Frozen DinoV2 ViT Small \\ \hline
\end{tabular}
\caption{\textbf{Soft Actor Critic Parameters.}}
\end{table}

\section{Additional Experimental Data}
\label{sup_additional_experiments}
\subsection{Run Visualization for Fold Cloth and Wipe Counter}
We include here the average success rates across the last 20 episodes for each of the three runs of each method in the ``Fold Cloth'' and ``Wipe Counter'' tasks. Training runs in the ``Wipe Counter'' task are less stable due to the base policy training on 10 demos being sensitive to reset conditions.
\begin{figure}[h] 
    \centering
    \includegraphics[width=0.99\linewidth, trim={0cm 0cm 0cm 0cm},clip]{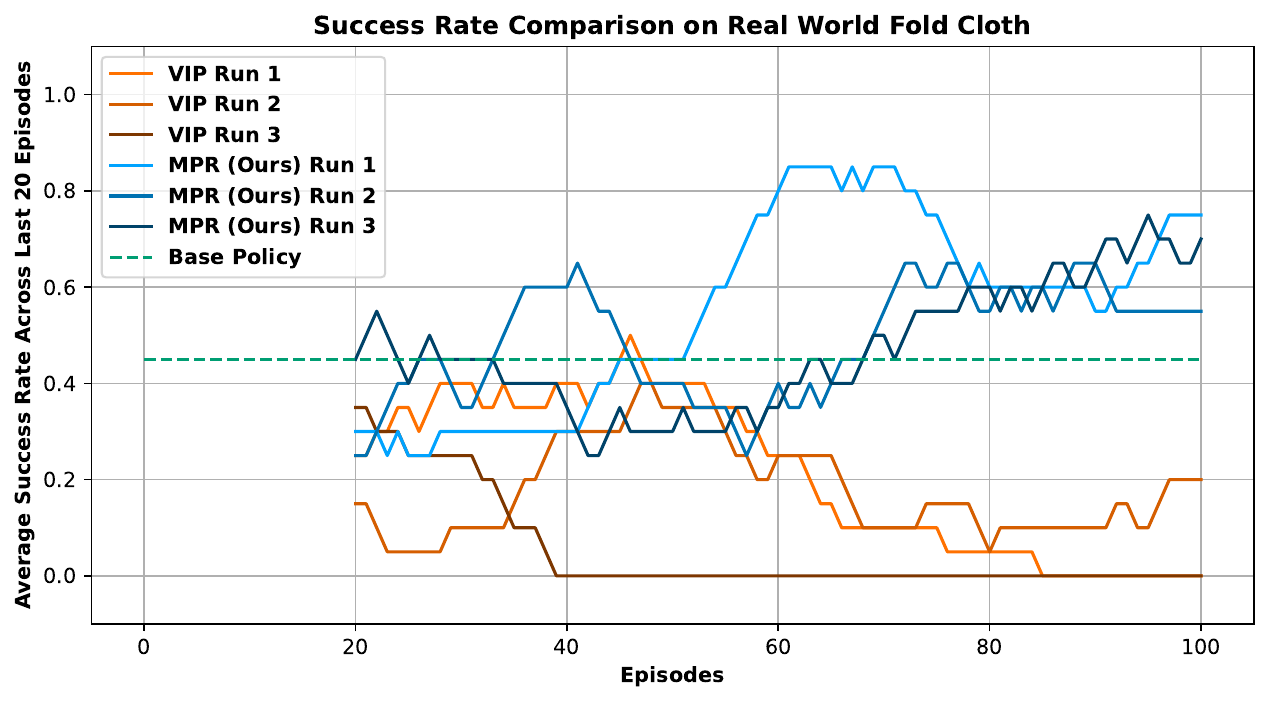}
  \caption{\textbf{Real-world training performance for the ``Fold Cloth'' task across three runs each for VIP \citep{maVIPUniversalVisual2023} and our approach, Motion Prediction Reward (MPR).} All policies were initialized with the same base policy capable of completing the task 45\% of the time (9/20 attempts) and trained for 100 episodes (about an hour of wall clock time).}
  \label{fig:Cloth_Training} 
  \vspace{-3mm}
\end{figure}
\begin{figure}[h] 
    \centering
    \includegraphics[width=0.99\linewidth, trim={0cm 0cm 0cm 0cm},clip]{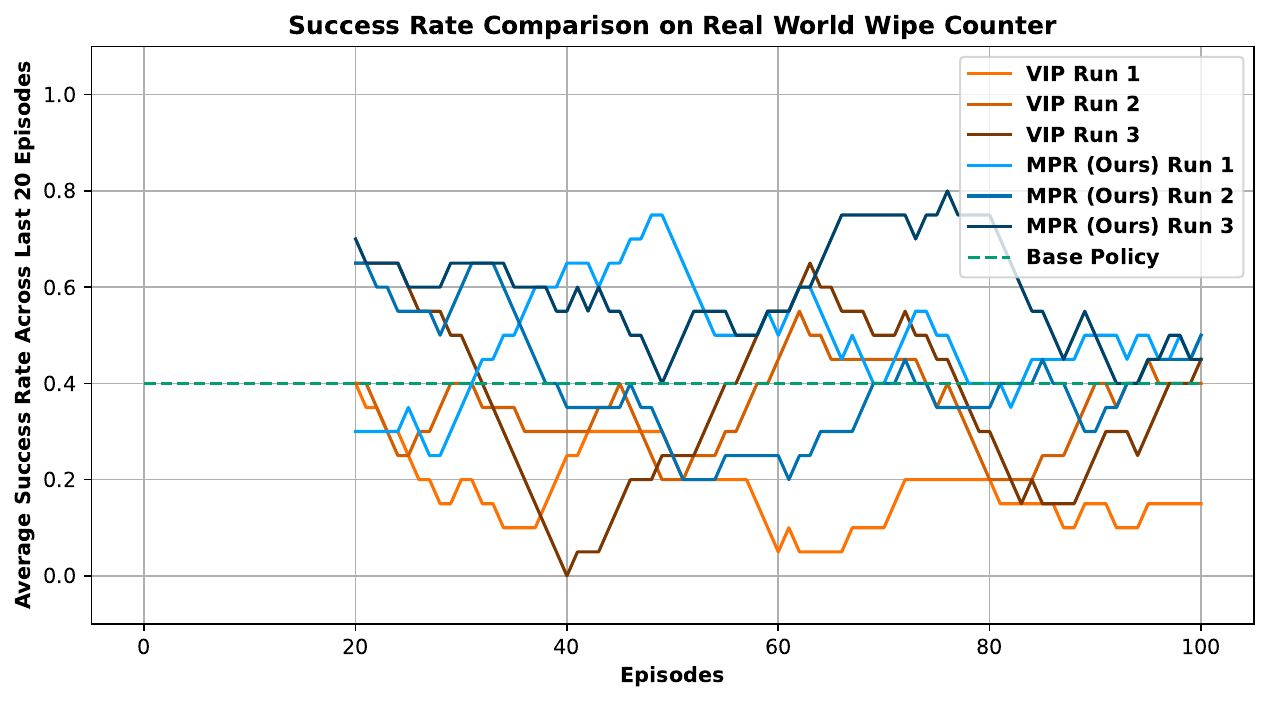}
  \caption{\textbf{Real-world training performance for the ``Wipe Counter'' task across three runs each for VIP \citep{maVIPUniversalVisual2023} and our approach, Motion Prediction Reward (MPR).} All policies were initialized with the same base policy capable of completing the task 40\% of the time (8/20 attempts) and trained for 100 episodes (about an hour of wall clock time).}
  \label{fig:Counter_Training} 
  \vspace{-3mm}
\end{figure}

\subsection{Ablation on Object Masking}
\begin{figure}[h] 
    \centering
    \includegraphics[width=0.99\linewidth, trim={0cm 0cm 0cm 0cm},clip]{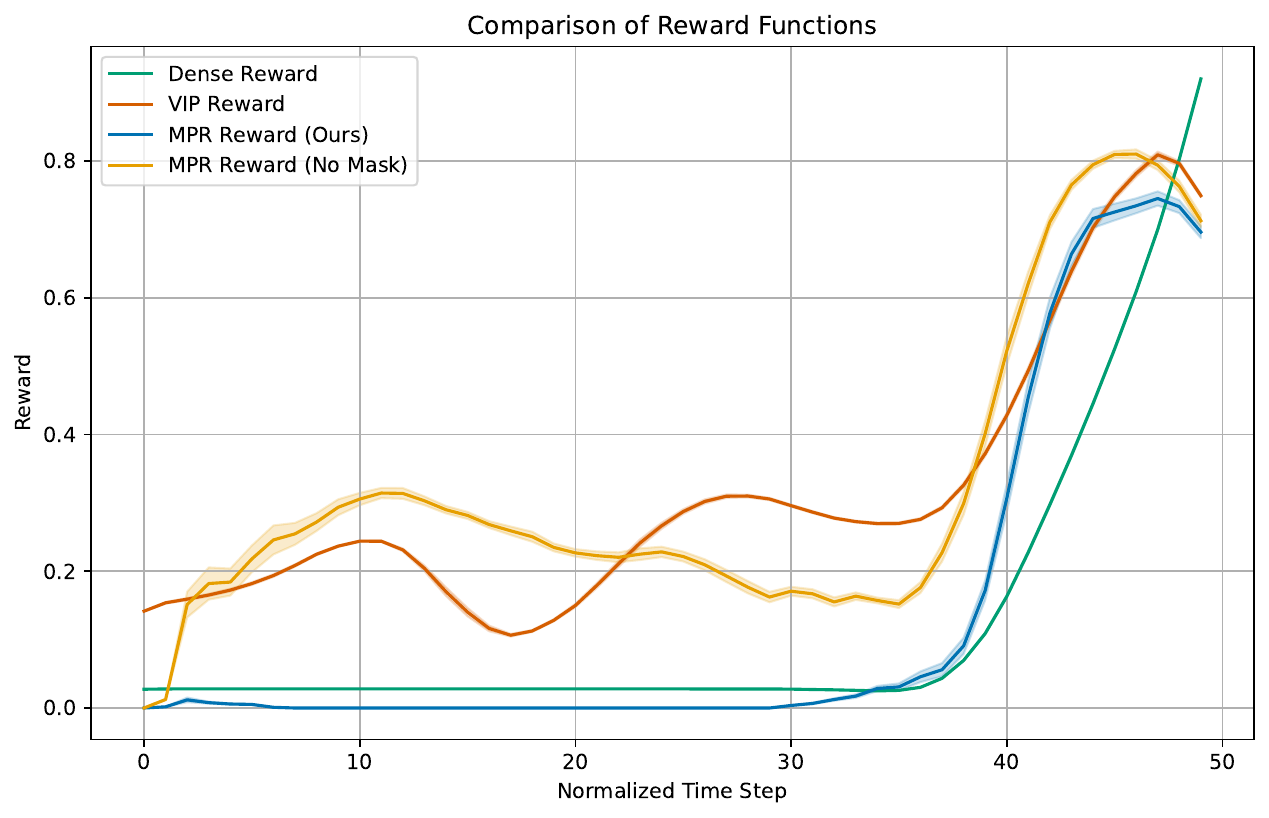}
  \caption{\textbf{Comparison of reward signals across 50 successful demonstrations for the simulated ``Open Microwave" task.} This plot includes an ablated version of our method that doesn't use task-relevant object masking.}
  \label{fig:Sim_Rewards_Ablation} 
  \vspace{-3mm}
\end{figure}

\begin{figure}[h] 
    \centering
    \includegraphics[width=0.99\linewidth, trim={0cm 0cm 0cm 0cm},clip]{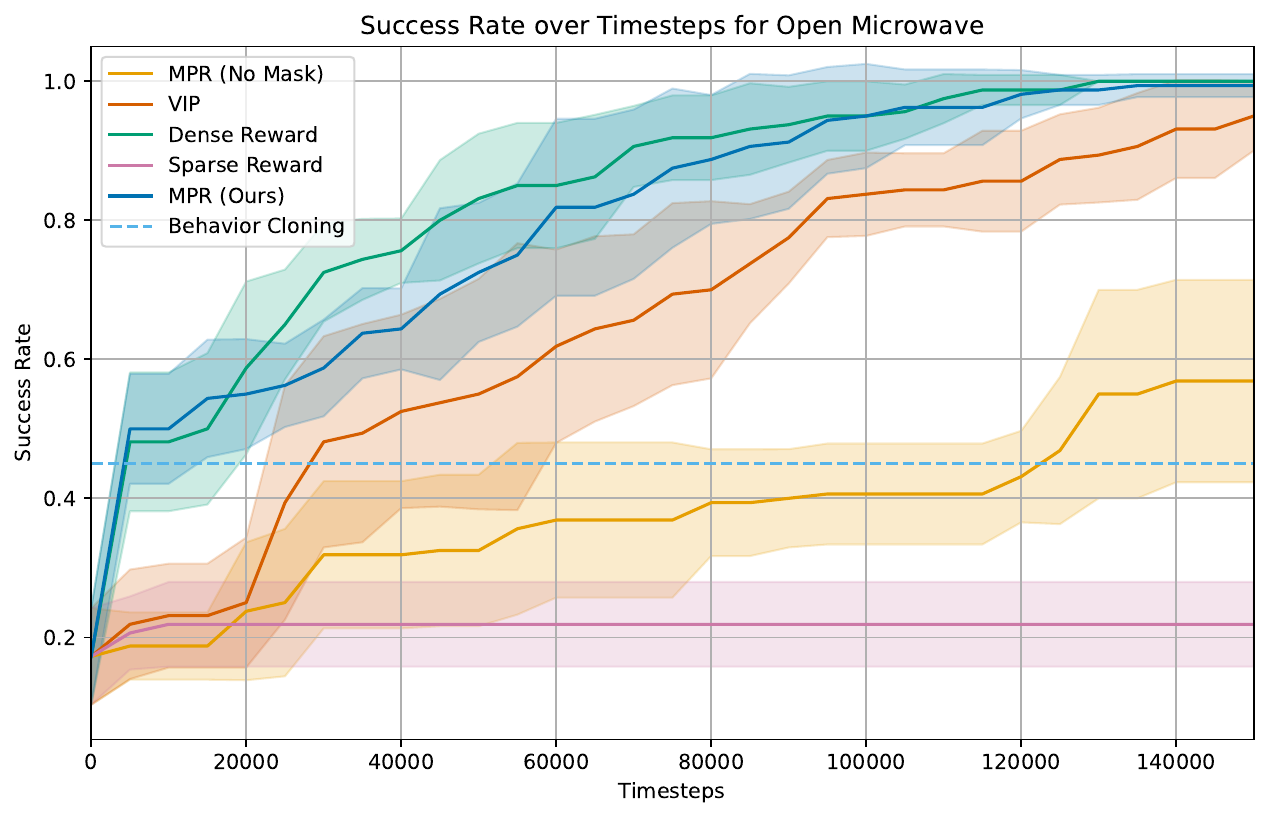}
  \caption{\textbf{Success rates for an ablated version of our Motion Prediction Reward without object masking evaluated on the open microwave task from the Franka Kitchen benchmark~\citep{guptaRelayPolicyLearning2019}.} The ablated version is less sample efficient.}
  \label{fig:Mask_Ablation} 
  \vspace{-3mm}
\end{figure}

We test the effects of removing the object-masking component from our training and inference pipelines. Instead of training to predict task-relevant object motion, we predict the motion of every point in the scene. This also removes our ability to subtract background motion from point tracks to compensate for camera motion in egocentric video. During training, this model exhibits overfitting behavior as evidenced by a rise in validation loss, which we didn't observe when training models with object masking. Figure \ref{fig:Sim_Rewards_Ablation} shows the reward signal on the open microwave task for this ablated version of the model, and Figure \ref{fig:Mask_Ablation} shows the learning performance during training. Without object masking, the learned reward signal is significantly noisier and results in less sample-efficient learning in the simulated task.


\subsection{Comparison to RoboReward}
We also evaluate our approach against RoboReward~\citep{leeRoboRewardGeneralPurposeVisionLanguage2026}, a robot reward foundation model. RoboReward is a Qwen3-VL 8 billion parameter model that has been finetuned on the Open X-Embodiment dataset annotated with reward values. RoboReward takes as input a video of task execution and the task language description, and outputs a sparse reward signal from 1-5 that rates the robot's performance on the task. While RoboReward outperforms the sparse reward signal provided by the environment (one for task success and zero otherwise), it underperforms all other methods that provide dense, per-step rewards.

\begin{figure}[h] 
    \centering
    \includegraphics[width=0.99\linewidth, trim={0cm 0cm 0cm 0cm},clip]{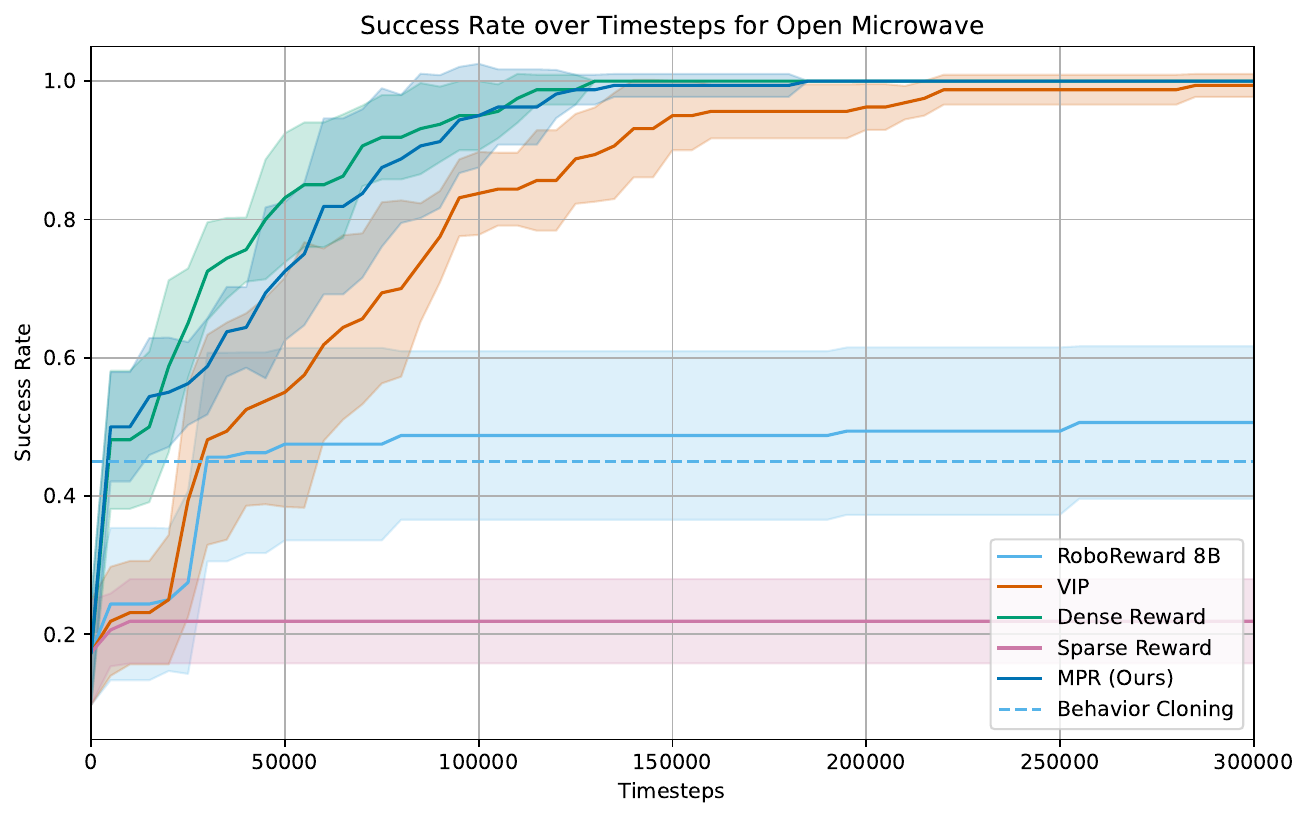}
  \caption{\textbf{Success rates including results with the RoboReward~\citep{leeRoboRewardGeneralPurposeVisionLanguage2026} 8 billion parameter model on the open microwave task from the Franka Kitchen benchmark~\citep{guptaRelayPolicyLearning2019}.}}
  \label{fig:RoboReward} 
  \vspace{-3mm}
\end{figure}

\subsection{Task Visualization}
 \begin{figure*}[t] 
    \centering
  \subfloat{%
       \includegraphics[width=0.20\linewidth,bb=0 0 640 480,trim={0cm 0cm 0cm 0cm},clip]{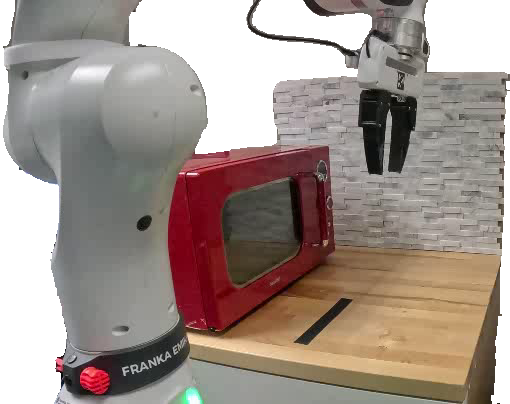}}
  \subfloat{%
        \includegraphics[width=0.20\linewidth,bb=0 0 640 480,trim={0cm 0cm 0cm 0cm},clip]{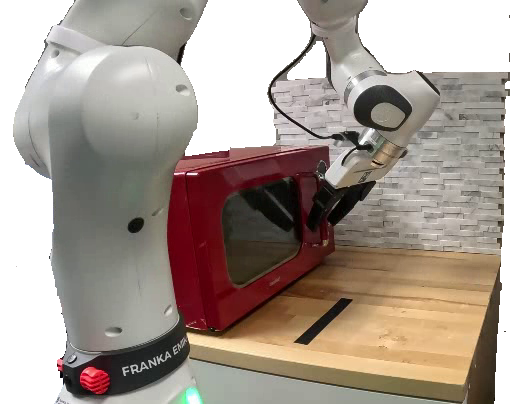}}
  \subfloat{%
        \includegraphics[width=0.20\linewidth,bb=0 0 640 480,trim={0cm 0cm 0cm 0cm},clip]{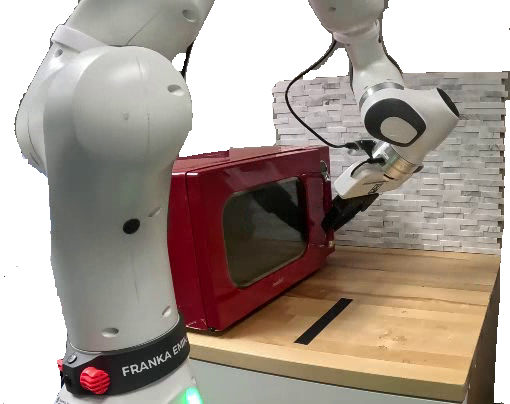}}
  \subfloat{%
        \includegraphics[width=0.20\linewidth,bb=0 0 640 480,trim={0cm 0cm 0cm 0cm},clip]{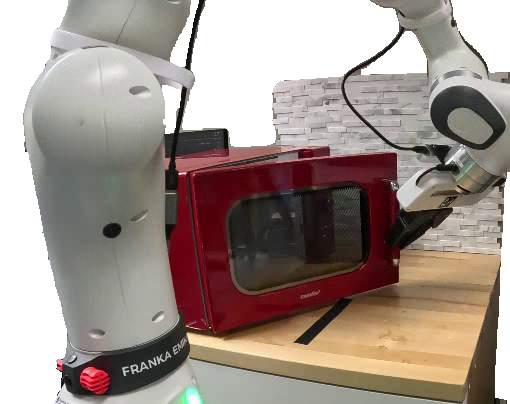}}
  \subfloat{%
        \includegraphics[width=0.20\linewidth,bb=0 0 640 480,trim={0cm 0cm 0cm 0cm},clip]{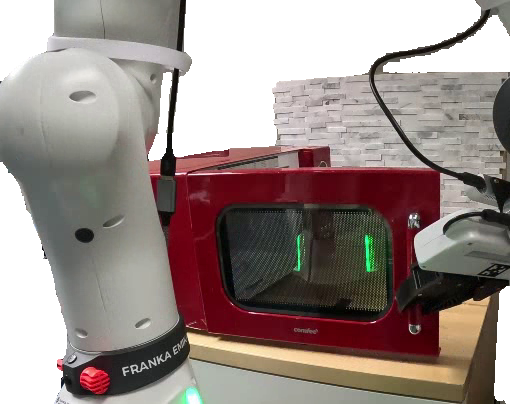}}\\
        \vspace{-4mm}
  \subfloat{%
       \includegraphics[width=0.20\linewidth,bb=0 0 640 480,trim={0cm 0cm 0cm 0cm},clip]{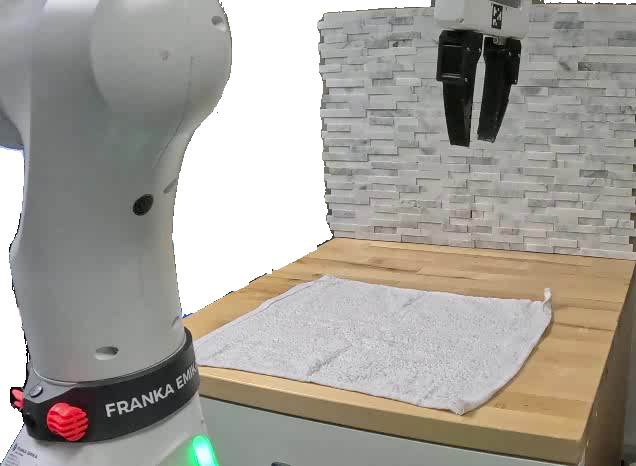}}
  \subfloat{%
        \includegraphics[width=0.20\linewidth,bb=0 0 640 480,trim={0cm 0cm 0cm 0cm},clip]{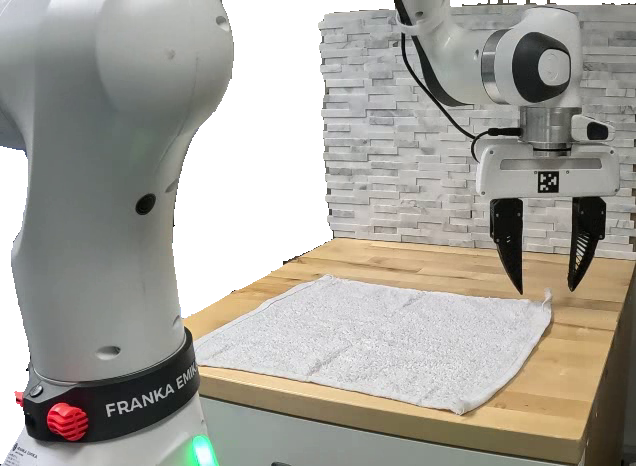}}
  \subfloat{%
        \includegraphics[width=0.20\linewidth,bb=0 0 640 480,trim={0cm 0cm 0cm 0cm},clip]{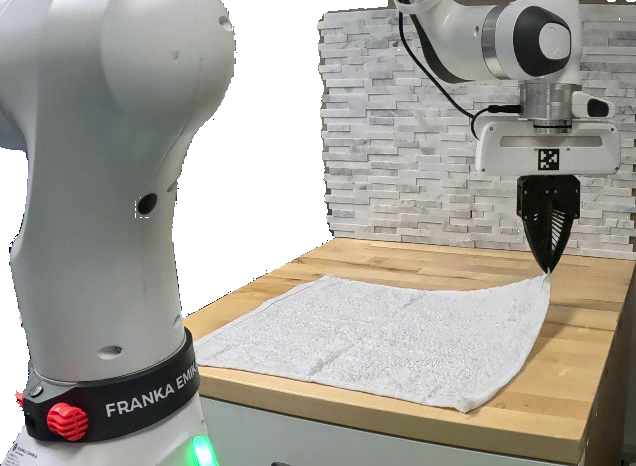}}
  \subfloat{%
        \includegraphics[width=0.20\linewidth,bb=0 0 640 480,trim={0cm 0cm 0cm 0cm},clip]{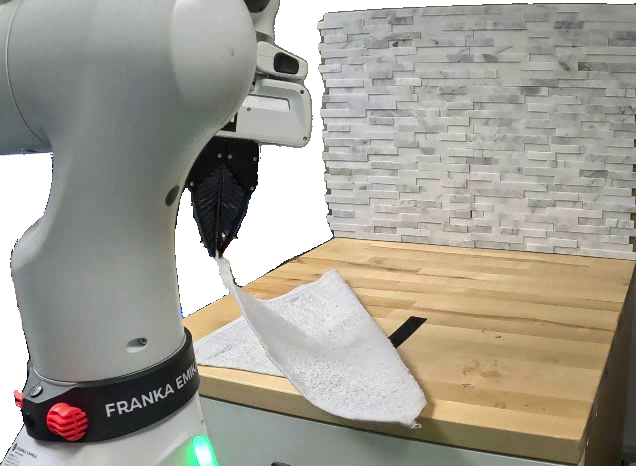}
        \includegraphics[width=0.20\linewidth,bb=0 0 640 480,trim={0cm 0cm 0cm 0cm},clip]{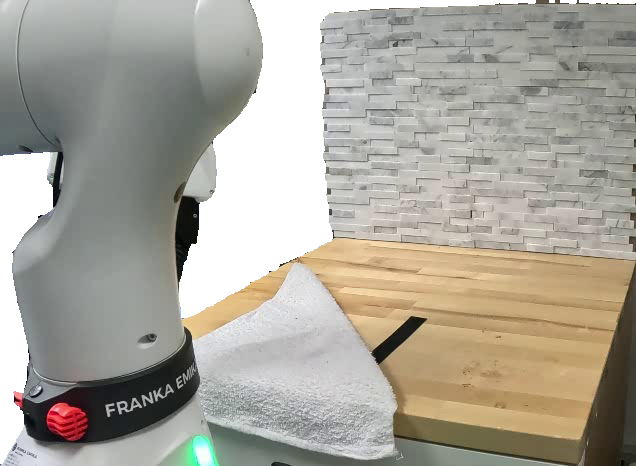}}\\
        \vspace{-4mm}
  \subfloat{%
       \includegraphics[width=0.20\linewidth,bb=0 0 640 480,trim={0cm 0cm 0cm 0cm},clip]{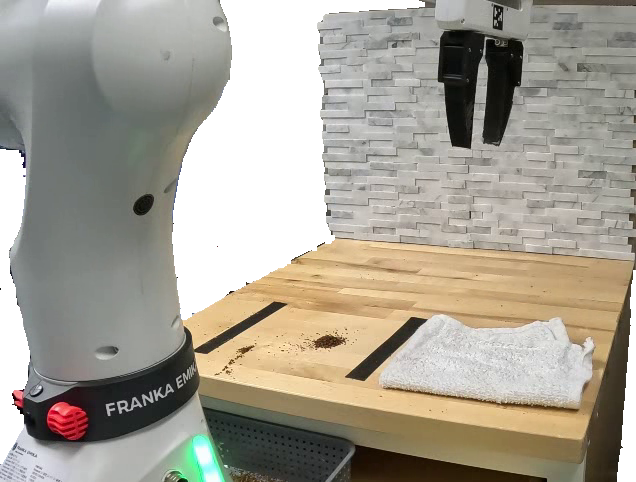}}
  \subfloat{%
        \includegraphics[width=0.20\linewidth,bb=0 0 640 480,trim={0cm 0cm 0cm 0cm},clip]{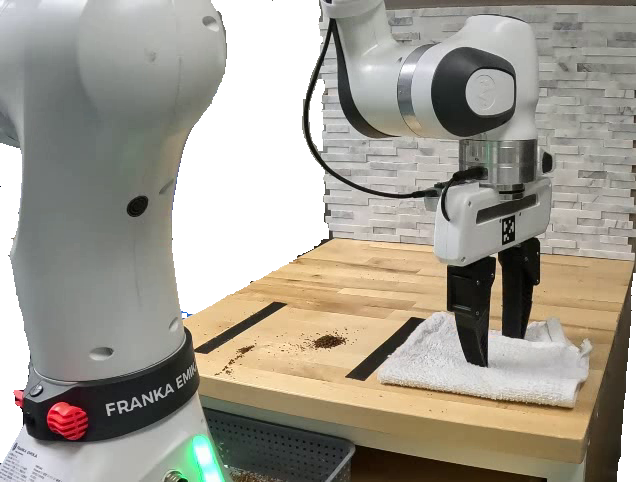}}
  \subfloat{%
        \includegraphics[width=0.20\linewidth,bb=0 0 640 480,trim={0cm 0cm 0cm 0cm},clip]{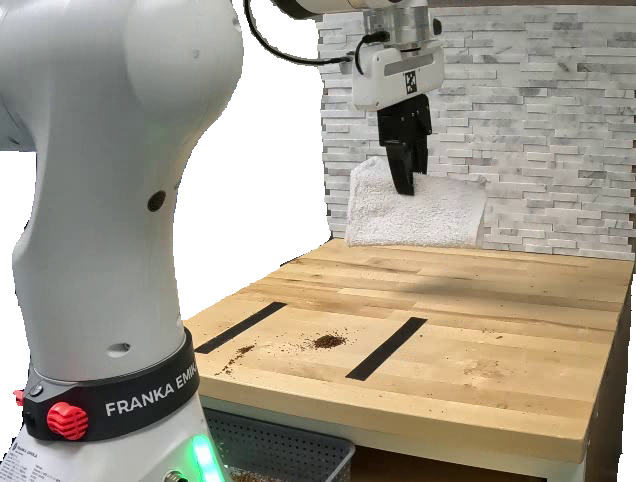}}
  \subfloat{%
        \includegraphics[width=0.20\linewidth,bb=0 0 640 480,trim={0cm 0cm 0cm 0cm},clip]{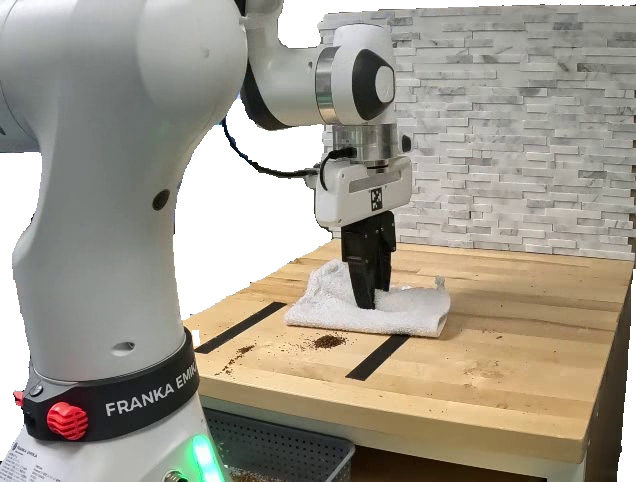}}
   \subfloat{%
        \includegraphics[width=0.20\linewidth,bb=0 0 640 480,trim={0cm 0cm 0cm 0cm},clip]{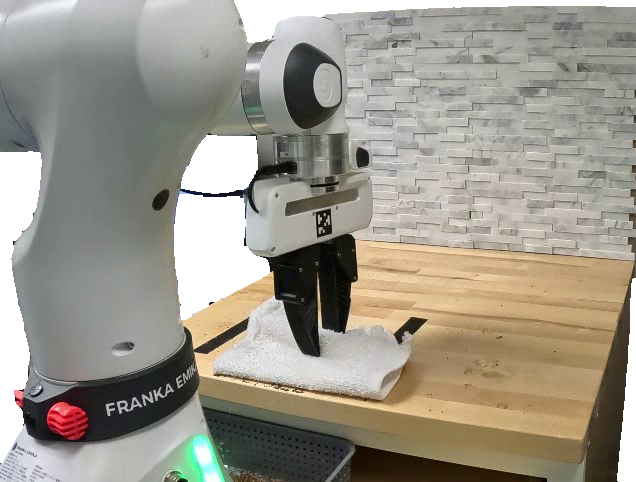}}\\

  \caption{\textbf{Visualization of the three different skills tested on hardware: ``Open Microwave'', ``Fold Cloth'', and ''Wipe Counter''.}}
  \label{fig:vis} 
   \vspace{-3mm}
\end{figure*}
Figure \ref{fig:vis} shows a visualization of the three tasks we tested on hardware.

\end{document}